\documentclass[sigconf]{acmart}

\usepackage{cleveref}
\usepackage{enumitem}
\usepackage{bm}

\usepackage{soul}
\newcommand{\cal}[1]{\mathcal{#1}}
\usepackage{color}
\definecolor{Orange}{rgb}{0.9,0.5,0}
\definecolor{NavyBlue}{rgb}{0.1, 0.4, 0.8}
\definecolor{Magenta}{rgb}{0.8, 0.1, 0.6}
\definecolor{mypink2}{RGB}{219, 48, 122}

\newcommand{\vpara}[1]{\vspace{0.05in}\noindent\textbf{#1}}

\usepackage{tabularx}
\usepackage{multirow}

\AtBeginDocument{%
  \providecommand\BibTeX{{%
    \normalfont B\kern-0.5em{\scshape i\kern-0.25em b}\kern-0.8em\TeX}}}

\copyrightyear{2020}
\acmYear{2020}
\setcopyright{acmlicensed}\acmConference[MM '20]{Proceedings of the 28th ACM International Conference on Multimedia}{October 12--16, 2020}{Seattle, WA, USA}
\acmBooktitle{Proceedings of the 28th ACM International Conference on Multimedia (MM '20), October 12--16, 2020, Seattle, WA, USA}
\acmPrice{15.00}
\acmDOI{10.1145/3394171.3413785}
\acmISBN{978-1-4503-7988-5/20/10}

\begin{document}
\fancyhead{}

\title{Retrieval Guided Unsupervised Multi-domain \\ Image-to-Image Translation}

\author{Raul Gomez}
\authornote{Both authors contributed equally to this research.}
\affiliation{%
	\institution{Centre Tecnol\`ogic de Catalunya}}
\affiliation{%
	\institution{Computer Vision Center, Spain}}
\email{raul.gomez@cvc.uab.es}

\author{Yahui Liu}
\authornotemark[1]
\affiliation{%
	\institution{University of Trento, Italy}}
\affiliation{\institution{Fondazione Bruno Kessler, Italy}}
\email{yahui.liu@unitn.it}

\author{Marco De Nadai}
\affiliation{\institution{Fondazione Bruno Kessler, Italy}}
\email{denadai@fbk.eu}

\author{Dimosthenis Karatzas}
\affiliation{%
	\institution{Universitat Aut\`onoma de Barcelona}}
\affiliation{%
	\institution{Computer Vision Center, Spain}}
\email{dimos@cvc.uab.es}

\author{Bruno Lepri}
\affiliation{\institution{Fondazione Bruno Kessler, Italy}}
\email{lepri@fbk.eu}

\author{Nicu Sebe}
\affiliation{%
	\institution{University of Trento, Italy}}
\affiliation{\institution{Huawei Research, Ireland}}
\email{niculae.sebe@unitn.it}

\renewcommand{\shortauthors}{Gomez, et al.}
\renewcommand{\shorttitle}{Retrieval Guided I2I Translation}

\begin{abstract}
Image to image translation aims to learn a mapping that transforms an image from one visual domain to another. 
Recent works assume that images descriptors can be disentangled into a domain-invariant content representation and a domain-specific style representation. 
Thus, translation models seek to preserve the content of source images while changing the style to a target visual domain.
However, synthesizing new images is extremely challenging especially in multi-domain translations, as the network has to compose content and style to generate reliable and diverse images in multiple domains.
In this paper we propose the use of an image retrieval system to assist the image-to-image translation task. First, we train an image-to-image translation model to map images to multiple domains. Then, we train an image retrieval model using real and generated images to find images similar to a query one in content but in a different domain. 
Finally, we exploit the image retrieval system to fine-tune the image-to-image translation model and generate higher quality images.
Our experiments show the effectiveness of the proposed solution and highlight the contribution of the retrieval network, which can benefit from additional unlabeled data and help image-to-image translation models in the presence of scarce data.

\end{abstract}



\maketitle

\section{Introduction}
Humans are remarkably good at generalizing. For example, given a picture of a blonde woman we can easily imagine in our mind how she would look like with black hair, even if we have never seen her with that look.  
Great efforts have been made to mimic this ability with learning-based models, especially through Generative Adversarial Networks (GANs)~\cite{goodfellow2014generative}. In particular, unsupervised image-to-image translation models focus on transferring visual appearance from one domain (e.g. blonde hair people) to another (e.g. black hair people) learning from unpaired images (i.e. no ground truth of the transformation is available)~\cite{zhu2017unpaired, liu2017unsupervised, huang2018multimodal, liu2020gmm, choi2018stargan}. However, despite the recent improvements in both quality and diversity of the results, it is still challenging to learn realistic transformations, especially with a limited amount of data.

Most of existing unsupervised models assume that all images can be grouped in visually distinctive categories (i.e. domains) having a domain-specific style and a domain-invariant content~\cite{huang2018multimodal, liu2020gmm, gonzalez2018image}. Thus, they try learning the latent spaces containing the content and the styles to synthetize images with the content of a source image and the style of a desired domain~\cite{zhu2017unpaired, huang2018multimodal, liu2020gmm, choi2018stargan}.
As there is an infinite number of mappings between two unpaired images, these approaches require a large amount of data.
This requirement is exacerbated in multi-domain problems, where a single model can map many domains. Despite recent results on few-shot transformations~\cite{liu2019few, anokhin2020high}, learning to generate visually realistic images in multiple domains and with limited data is still an open problem.

In this paper, we present a novel approach that helps to learn better mappings between domains. To do so, we propose the use of an image Retrieval model to Guide and improve UNsupervised Image-to-image Translation performance, i.e., RG-UNIT. 
Our model consists of three parts. First, an image-to-image translation model that learns to translate images to multiple domains by disentangling style and content. Second, an image retrieval system that exploits the image-to-image translation model to learn to find images similar to the source one, but in the target visual domain.
Third and final, we fine-tune the image-to-image translation model exploiting the information provided by the image retrieval system, which results in a performance improvement.

The contributions of this paper are as follows:
\begin{itemize}
    \item We propose the use of a retrieval system to improve image-to-image translation models at combining content and style latent spaces and thus generate higher quality images from multi-modal and multi-domain image-to-image translations;
    \item To our knowledge, we are the first to train a retrieval system exploiting image-to-image translation model generated images and then use it to fine-tune the image translation model, a technique that can take advantage of unlabeled data;
    \item We validate the proposed solution in the challenging task of multi-domain translation of facial attributes on CelebA~\cite{liu2015deep} dataset, where style is defined as a combination of attributes. 
    Quantitative and qualitative results show better performance in all experiments, especially when additional unlabeled data is exploited.
\end{itemize} 
\vspace{-1em}

\section{Related Work}
In this paper we leverage an image retrieval system to ease the difficult task of translating an image from one domain (e.g. blonde hair people) to another one (e.g. black hair people).
Thus, our work is at the meeting point between image-to-image translation and image retrieval.

\vpara{Image Translation.} Image-to-image translation models usually generate new images through conditional Generative Adversarial Networks (cGANs)~\cite{goodfellow2014generative, mirza2014conditional}, where an adversarial game between a generator and a discriminator helps learning to synthetize images in the target domain.
Isola \emph{et al.}~\cite{isola2017image} proposed an encoder-decoder network that tries to reconstruct the source image in the target domain, instead of reconstructing the source image in the original one. 
Wang \emph{et al.}~\cite{wang2018high} later extended it to higher resolutions. A desirable property of image-to-image translation is the ability to generate multiple plausible translations~\cite{gonzalez2018image}, as there are many different translations of an image from one domain to another. For example, a nightlight picture might be translated to multiple daylight pictures with different weather/light conditions. Hence, Zhu \emph{et al.}~\cite{zhu2017toward} mixed GANs and Variational Auto-Encoders (VAEs) to learn a latent distribution and later sample from this distribution. 
However, it is expensive to collect pairs of images in different domains (e.g. daylight $\leftrightarrow$ nightlight scene with people in the same pose) and often impossible (e.g. a picture of the author of this paper 10 years from now).
Thus, the most interesting results come undoubtedly from unsupervised image-to-image translation which learns a mapping from unpaired images (e.g. daylight $\leftrightarrow$ nightlight pictures of different places).
Liu \emph{et al.}~\cite{liu2017unsupervised} assumed that all images share a domain-invariant latent space and force all images to map into it before translating them to the target domain.
Zhu \emph{et al.}~\cite{zhu2017unpaired} outperformed previous results with CycleGAN, a network that requires generated images to be translated back in the original domain after the translation.
Mo \emph{et al.}~\cite{mo2018instanceaware} extended it to multiple instances per image, while Huang \emph{et al.}~\cite{huang2018multimodal} assumed that domains share a common content space but different style spaces, which help achieving multi-modality in an unsupervised setting.
Recently, literature extended the previous models~\cite{choi2018stargan, liu2020gmm} by modelling multiple domains at once (e.g. blonde person to be translated to black, brown hair colours). 
Choi \emph{et al.}~\cite{choi2018stargan} proposed StarGAN which uses a domain label and a domain classifier to map images in different domains.
Liu \emph{et al.}~\cite{liu2020gmm} later proposed a unifying approach to multi-modal translation in multiple domains by using a VAE-like approach to model the latent styles of the domains with a Gaussian Mixture Model (GMM).

\vpara{Image Retrieval.}
Image retrieval is the task of finding images similar to a given query. In the image-to-image scenario the query is composed by an image and distances between images in an embedding space are optimized using a triplet ranking loss~\cite{WangTriplet,SchroffTriplet}.
However, queries are not necessarily limited to images. 
Considerable effort has been made especially in using textual queries~\cite{Karpathy,Kiros2014,Wang,Patel}, where researchers focus on learning alignments between images and textual representations.
Others exploited image labels~\cite{Wu, Satpute} such as Siddiquie \textit{et al.}~\cite{Siddiquie2011} who used multi-attribute queries, modeling attributes correlations, and Scheirer \textit{et al.}~\cite{Scheirer} focused on attribute-based visual similarity searches.

Queries might also be multi-modal, as in the case of finding images similar to a given picture but with some characteristics specified by a condition (usually a textual sentence).
Smith \textit{et al.}~\cite{Smith} proposed an image-to-image retrieval setup where image queries are graphically modified by user-mouse interaction.
Gordo \textit{et al.}~\cite{DianeLarlus2017} used textual captions associated with images to learn a shared embedding space for images and text. Gomez \textit{et al.}~\cite{Gomez2019} instead learned this embedding space from social media data in a weakly-supervised setup.

In this paper we aim to retrieve images relevant to complex queries composed by image content and image style encodings, with the peculiarity that those representations are extracted by pre-trained feature extractors and from synthetically generated images. 
In our experiments we work with face images, were style is defined by face attributes (i.e blond hair) and the content is the visual appearance not defined by those attributes.
As far as we know, we are the first ones proposing that setup for a retrieval model training, and using it later to improve the image translation model performance.
The recent work of Anoosheh \emph{et al.}~\cite{anoosheh2019night} is the only similar approach we know about. They propose to localize previously seen places through image retrieval systems but, instead of exploiting a large amount of labeled data, they use an image translation model to generate images of places under different weather conditions, and therefore enlarge their dataset.
Unlike the mentioned method, we learn image representations in a multi-domain setting, exploring different image translation strategies, and we use the learned image retrieval model to improve the image translation performance.

\begin{figure*}[ht]
	\centering
  \includegraphics[width=0.8\linewidth]{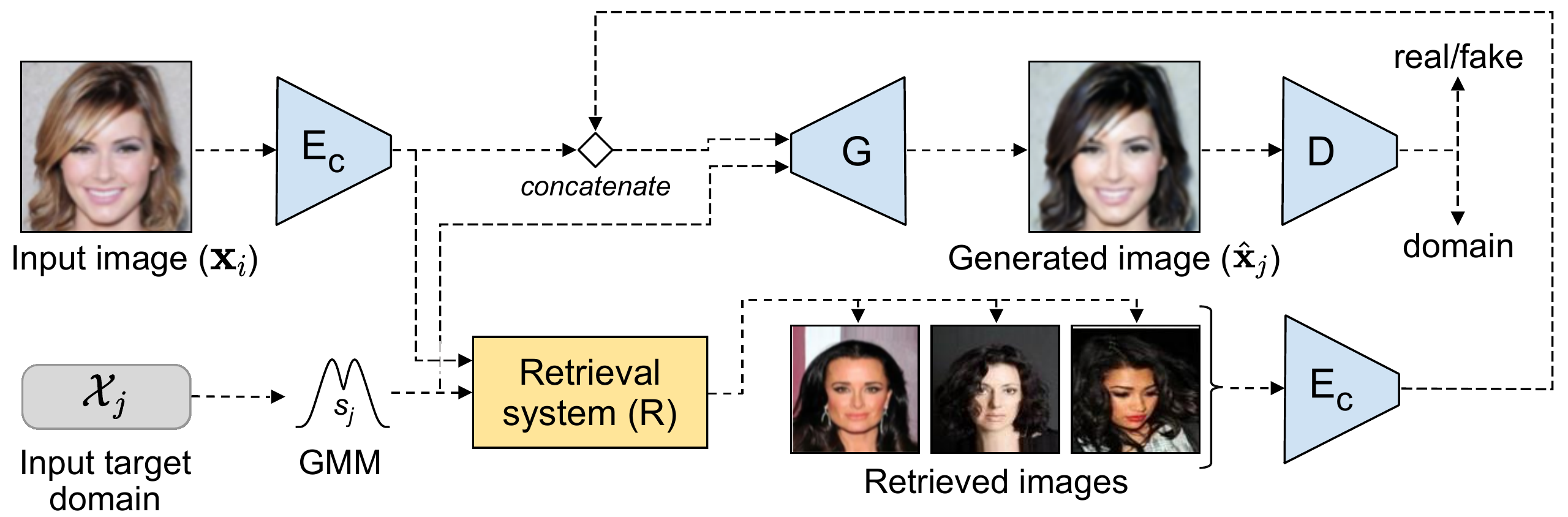}
   \caption{RG-UNIT training pipeline. We here translate an input image $\mathbf{x}_i \in \mathcal{X}_i$ to domain $\mathcal{X}_j$. We sample a style $\mathbf{s}_j$ from the latent GMM distribution, then we feed the generator with $\mathbf{s}_j$ and we merge with the content features of the input image ($E_c(\mathbf{x}_i)$) with the content features of the retrieved images.}
   \label{fig:RG_pipeline}
\end{figure*}

\section{Our Method}
Given a collection of image domains $\{\mathcal{X}_1, \mathcal{X}_2, \cdots, \mathcal{X}_K\}$, we aim at transforming an input image $\bm{x}\in\mathbb{R}^{H\times W\times 3}$ belonging in domain $\mathcal{X}_i$ to domain $\mathcal{X}_j$ without requiring paired images (here $i, j$ refer to domain indices). 
Following recent seminal work~\cite{lee2018diverse,huang2018multimodal,liu2020gmm}, we assume that each image $\bm{x}$ shares a domain-invariant content latent representation $c \in \mathcal{C}$ and has a domain-specific latent representation (also called style) $\bm{s}_i \in \mathcal{S}_i$, where $\mathcal{S}$ is termed as the style latent space.
Thus, for each image $\bm{x} \in \mathcal{X}_i$, we would like to generate an image $\tilde{\bm{x}}\in\mathcal{X}_j$ with the same content as $\bm{x}$  (e.g., a person looking left) but the style of $\mathcal{X}_j$ (e.g., blonde hair).

Following the general framework GMM-UNIT~\cite{liu2020gmm}, we model the style latent space through a $K$-component $d$-dimensional Gaussian Mixture Model (GMM). Formally, its probability density $p(\mathbf{s})$ is defined as: 
\begin{equation}
    p(\mathbf{s}) = \sum_{k=1}^{K}\phi_k\mathcal{N}(\mathbf{s}; \pmb{\mu}_k, \pmb{\Sigma}_k)
\end{equation} 
where $\phi_k$ denotes the weight ($\phi_k\geq 0$, $\sum_{k=1}^K \phi_k = 1$), $\pmb{\mu}_k \in \mathbb{R}^d$ and $\pmb{\Sigma}_k\in\mathbb{R}^{d\times d}$ refer to the mean and the covariance matrix of the Gaussian component $k$, respectively.
Thus, each domain $\mathcal{X}_i$ is represented by the $i$-th GMM component $\bm{s}_i\sim{\cal N}(\pmb{\mu}_i, \pmb{\Sigma}_i)$. 
The content latent space is instead modeled through the network, which is forced to understand the domain-invariant part of images. 

We propose to use an image retrieval system to assist the network on this difficult task.
Figure~\ref{fig:RG_pipeline} depicts our architecture. The model is trained in an adversarial fashion and it's composed of a content encoder $E_c$, which extracts the content information from an image, a latent GMM distribution where we sample the style in a Variational Auto-Encoder (VAE) fashion, a retrieval system $R$, a generator $G$ and a discriminator $D$. The network also has a style encoder $E_s$ (not shown in the picture), which extracts the style from an image by predicting the parameters of the GMM. $E_s$ will be used to learn the latent distribution.

The inputs of the model are a source image and a target domain. During a forward pass, the retrieval system $R$ finds the $r$ most similar images in content, but with the style of the target domain. Then, the content extracted from these images is used along with the content extracted from the input image and the target style to synthesize a new image.
Our hypothesis is that the retrieved images can provide information to the decoder about how the input images should look like in the target domain, leading it to generate more realistic images.

Training together the retrieval system and the image translation models would be very challenging as we would need to jointly learn to disentangle content and style, generate new images and also retrieve the optimal images. Thus, we split the training in three stages. 
First, we train an image-to-image translation model that is not guided by retrieved images ($r=0$) following the GMM-UNIT network architecture~\cite{liu2020gmm}. Second, exploiting its generated images and its content and style features disentanglement, we train an image retrieval system.
Finally, we train the RG-UNIT image translation model with the retrieval model guidance. In this last training stage, we freeze the encoders $E_c$, $E_s$ because the retrieval model relies on them, so updating their weights would hurt the retrieval model performance.

\subsection{Translating images}
\label{sec:translating_images}
In this section we focus on the first training step of our method: The image-to-image translation model that does not rely on the image retrieval system.
Translating images from one domain to another without paired images is a very challenging task, since the model has to learn to disentangle the style and the content from unpaired input images (i.e. different people in two domains). 

Given an image $\bm{x}\in \mathcal{X}_i$ and a target image domain $\pmb{\mathcal{X}}_j$, let $\bm{s}\sim{\cal N}(\pmb{\mu}_j, \pmb{\Sigma}_j)$ be the latent style sampled from the GMM component.
Then, we extract the content from the input image $\bm{c}=E_c(\bm{x})$ and we use it together with $\bm{s}$ to synthesize a new image.
Since there are many combinations of content and style that result in the input image, the problem has to be constrained through different losses.

\medskip \textbf{Content and style disentanglement.} Similar to~\cite{huang2018multimodal, liu2020gmm}, we learn to disentangle content and style by requiring the content and style latent representations to be correctly reconstructed after the translation. So, we use a content encoder $E_c$ and a style encoder $E_s$ and impose:
\begin{equation}
    \mathcal{L}_{\textrm{c/recon}} = \textstyle \mathbb{E}_{\bm{x}, \bm{s}} \big[\| E_c(G(\tilde{\bm{c}}, \bm{s})) - E_c(\bm{x})\|_1 \big]
    \label{eq:1}
\end{equation}
\begin{equation}
    \mathcal{L}_{\textrm{s/recon}} = \textstyle \mathbb{E}_{\bm{x}, \bm{s}} \big[\| E_s(G(\tilde{\bm{c}}, \bm{s})) - E_s(\bm{x})\|_1 \big]
    \label{eq:2}
\end{equation}
where $E_s(\cdot)$ refers to the style extractor encoding images into style latent space, and $\tilde{\bm{c}}$ refers to the latent content. For the image-to-image translation model without retrieval $\tilde{\bm{c}} = E_c(\bm{x})$, we will modify this equivalence in \Cref{sec:retrievalTranslation} by leveraging the trained retrieval system.

\medskip \textbf{Pixel-level reconstruction.} To encourage the pixel-level consistency of generated and real images, we employ a reconstruction loss and a cycle consistency loss~\cite{almahairi2018augmented, zhu2017unpaired}, generating an image and then translating it back to the original one.
\begin{equation}
    \mathcal{L}_{\textrm{x/recon}} = \textstyle \mathbb{E}_{\bm{x}, \bm{s}} \big[\| G(\tilde{\bm{c}}, \bm{s}) - \bm{x}\|_1 \big]
    \label{eq:3}
\end{equation}
\begin{equation}
    \mathcal{L}_{\textrm{cycle}} = \textstyle \mathbb{E}_{\bm{x}, \bm{s}} \big[\| G(E_c(G(\tilde{\bm{c}}, \bm{s})), E_s(x)) - \bm{x}\|_1 \big]
    \label{eq:4}
\end{equation}
The $\mathcal{L}_1$ loss was found to encourage the generation of sharper images than the $\mathcal{L}_2$~\cite{isola2017image}.

\medskip \textbf{Latent distribution.} We aim to learn a style latent distribution from which we can sample a style code and translate one image from one domain to another. Inspired by the VAEs approaches, we encourage the encoder conditional distribution to match the prior latent distribution with the Kullback-Leibler divergence. 
Since the latent GMM we define is diagonal, it can be computed as:
    \begin{equation*}
        \mathcal{L}_{\textrm{KL}} = \textstyle \sum_{k=1}^K\mathbb{E}_{\bm{x}\sim p_{\pmb{\mathcal{X}}^k}} [\mathcal{D}_{KL}(E_s(\bm{x})\|\mathcal{N}(\pmb{\mu}_k, \pmb{\Sigma}_k))]
        \label{eq:kl}
    \end{equation*}
where $\mathcal{D}_{KL}(p\|q) = -\int p(t)\log\frac{p(t)}{q(t)}dt$ is the Kullback-Leibler divergence.
This loss ultimately is expected to learn the true posterior of the latent distribution.

\medskip \textbf{Domain loss.} To support the network at learning to generate images in the correct domain, we employ a classification loss on the discriminator~\cite{choi2018stargan}. Thus,
\begin{align*}
    \mathcal{L}_{\text{cls}}^D &= \textstyle \mathbb{E}_{\bm{x}, d_{\pmb{\mathcal{X}}_i}}[-\log D_{\text{cls}}(d_{\pmb{\mathcal{X}}_i}|\bm{x})],\\
    \mathcal{L}_{\text{cls}}^G &= \textstyle \mathbb{E}_{\bm{x},d_{\pmb{\mathcal{X}}_j}, \bm{s}
    }[-\log D_{\text{cls}}(d_{\pmb{\mathcal{X}}_j}|G(\tilde{\bm{c}}, \bm{s}))]
    \label{eq:5}
\end{align*}
where $d_{\pmb{\mathcal{X}}}$ is the label of domain. The generator and discriminator are trained using the $\mathcal{L}_{\text{cls}}^G$ and $\mathcal{L}_{\text{cls}}^D$ loss, respectively.

\begin{figure*}[!ht]
	\centering
  \includegraphics[width=0.7\linewidth]{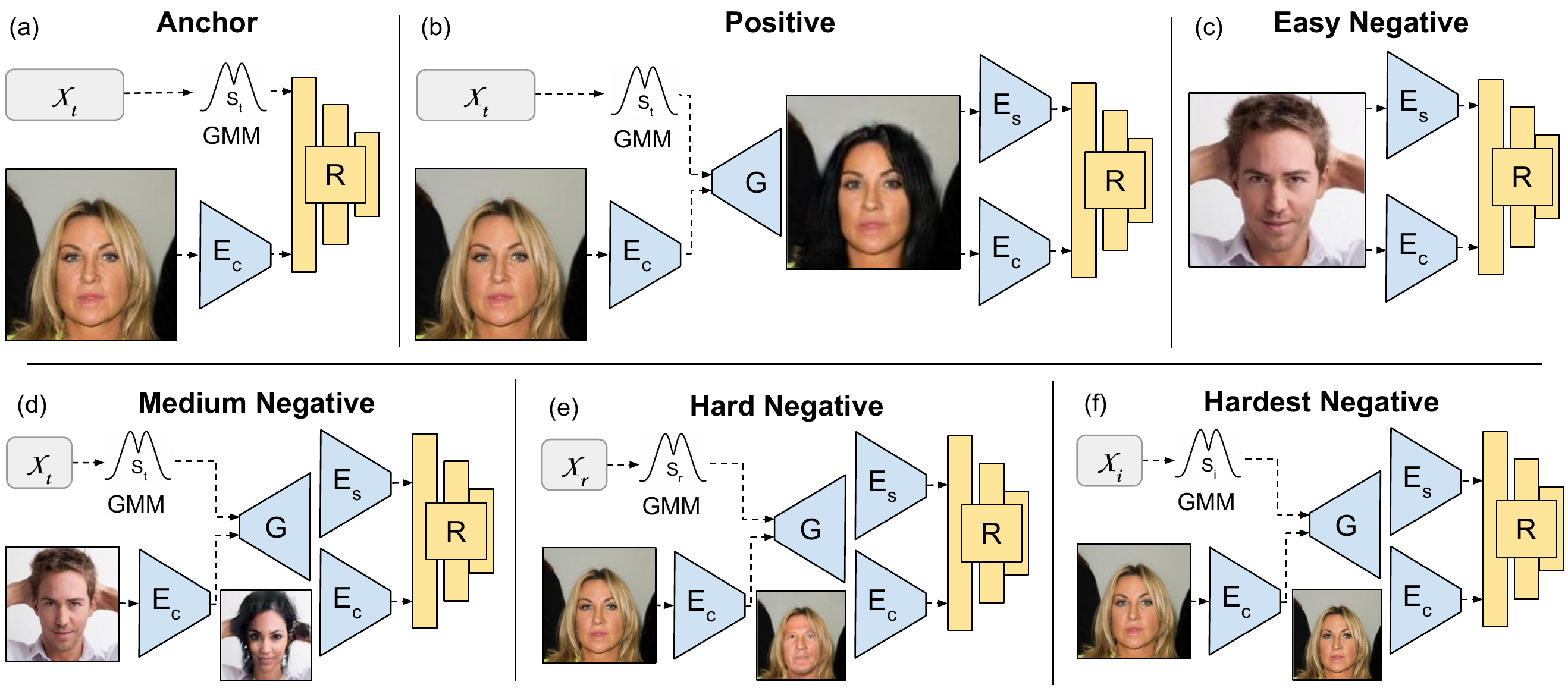}
   \caption{Retrieval model triplets creation pipeline. In this example, the target domain $t$ is: Black Hair, Female, Young. The domain $i$ is the one of the input image, and $r$ is a randomly selected domain different than $t$. The box R stands for the retrieval model, which outputs 300 dimensional vectors that are compared using a ranking loss.}
   \label{fig:R_pipeline}
\end{figure*}

\medskip \textbf{Adversarial game.} We train our model in an adversarial fashion with a Least Square loss~\cite{mao2017least}.
We construct the adversarial game with:
\begin{align}
\mathcal{L}_{\textrm{r/f}}^D = &\mathbb{E}_{\bm{x}}\bigl[(D_\text{ real}(\bm{x}))^2\bigr]+ \mathbb{E}_{\bm{x}, \bm{s}}\bigl[(D_\text{real}(G(\tilde{\bm{c}}, \bm{s}))-1)^2\bigr] \\
\mathcal{L}_{\textrm{r/f}}^G = &\mathbb{E}_{\bm{x},\bm{s}}\bigl[(D_\text{real}(G(\tilde{\bm{c}}, \bm{s})))^2\bigr]
\label{eq:6}
\end{align}
where $D_\text{real}$ is the discriminator predicting whether an image is real or fake.

The final optimization problem is defined as:
\begin{align*}
\mathcal{L}_D =& \mathcal{L}_{\textrm{r/f}}^D + \mathcal{L}_{\text{cls}}^D\\
    \mathcal{L}_G =& \mathcal{L}_{\textrm{r/f}}^G + \mathcal{L}_{\text{c/recon}} + \lambda_{\text{s/recon}} \mathcal{L}_{\text{s/recon}} +   \mathcal{L}_{\text{x/recon}} \\ &+ \lambda_{\text{cycle}}\mathcal{L}_{\text{cycle}} +  \lambda_{\text{KL}}\mathcal{L}_{\text{KL}} + 
    \mathcal{L}_{\text{dom}}^G 
\end{align*}
where $\{\lambda_{\textrm{s/rec}}, \lambda_{\textrm{cyc}}, \lambda_{\textrm{KL}}\}$ are hyper-parameters of weights for the corresponding loss terms. The values of these parameters come from the literature. We refer to the Supplementary material for details.

\subsection{Retrieval System}
\label{sec:retrieval}
The style and content disentanglement is particularly difficult and often requires multiple constraints and tricks~\cite{huang2018multimodal,lee2018diverse,liu2020gmm} to converge to a satisfactory result. 
To reduce the load on the generator, we use a retrieval system to fetch images similar in content to the input image but with the target style, which might help the decoder to generate more realistic images.


We train a network with three branches that share weights. Each one has as input content and style latent representations and outputs an embedded representation of the input information.
We train the network with a triplet ranking loss~\cite{chechik2010large} to produce close representations for elements with a similar content and in the same domain. 
Specifically, given triplet samples $\langle \bm{r}_a, \bm{r}_p, \bm{r}_n\rangle$ (i.e. anchor, positive and negative samples, respectively), we train the model to achieve that the Euclidean distance between the anchor and the negative samples is greater (and bigger than a margin) than the Euclidean distance between the anchor and the positive samples. We define the loss as:
\begin{equation}
    \mathcal{L}_{\textrm{ret}} = \max(0, m + d(\bm{r}_a,\bm{r}_p) - d(\bm{r}_a,\bm{r}_n))
\end{equation}
where $m>0$ is a constant margin.

We use a pre-trained image-to-image translation model to get disentangled representations of images content and style (GMM-UNIT in our experiments~\cite{liu2020gmm}), which are the ones fed to the proposed retrieval system. The image-to-image translation model is also used to generate images for the training triplets. As far as we know, we are the first ones to use image-to-image translation models features' disentanglement and generated images to train an image retrieval system.



Specifically, given an input image $\bm{x} \sim p_{\pmb{\mathcal{X}}_i}$ with content $\bm{c} = E_c(\bm{x})$ and a target style $\bm{s}_t\sim{\cal N}(\pmb{\mu}_t, \pmb{\Sigma}_t)$, we define an anchor sample as $\bm{r}_a = (\bm{c}, \bm{s}_t)$, as shown in~\Cref{fig:R_pipeline}(a).
To get the positive sample, we use the pre-trained translation network to generate an image $\hat{\bm{x}} = G(\bm{c}, \bm{s}_t)$  in the target domain and define the triplet positive sample as $\bm{r}_p = (E_c(\hat{\bm{x}}), E_s(\hat{\bm{x}}))$ (see~\Cref{fig:R_pipeline}(b)).
The negative samples $r_n$ are defined with different competing strategies that lead the network to learn distances between samples based both on content and attributes. Being $\bm{x}_r\in\mathcal{X}_r$ a random training image, where $t\neq r$, we define the following strategies for the negatives samples mining:

\begin{itemize}[leftmargin=*, topsep=4pt]
    \item \textit{Easy Negative}: $\bm{r}_n = (E_c(\bm{x}_r), E_s(\bm{x}_r))$. We use as negative the extracted content and style of the random image (see~\Cref{fig:R_pipeline}(c)).
    \item \textit{Medium Negative}: $r_n =  (E_c(\hat{\bm{x}}_r), E_s(\hat{\bm{x}}))$, where $\hat{\bm{x}}_r=G(E_c(\bm{x}_r), \bm{s}_t)$. We translate $\bm{x}_r$ to the target style $\mathcal{X}_t$ and use it as negative. These negative samples have the same style as the anchor and positive samples, but different content, as shown in~\Cref{fig:R_pipeline}(d). 
    \item \textit{Hard Negative}: $r_n =  (E_c(\hat{\bm{x}}_r), E_s(\hat{\bm{x}}_r))$, where $\bm{s}_r \in \mathcal{S}_r$ and $\hat{\bm{x}_r} = G(\bm{c}, \bm{s}_r)$. We translate the anchor content to a random style $\mathcal{S}_r$ and use it as negative. These negatives will have similar content as the anchor and positive samples, but different style (see~\Cref{fig:R_pipeline}(e)). We also create another hard negative: $r_n =  (E_c(\tilde{\bm{x}}), E_s(\tilde{\bm{x}}))$, where $\bm{s}_i \sim{\cal N}(\pmb{\mu}_i, \pmb{\Sigma}_i)$ and $\tilde{\bm{x}} = G(\bm{c}, \bm{s}_i)$. We sample a transformation of $\bm{x}$ in its original domain $\mathcal{X}_i$ and use it as negative. These negatives will have similar content as the anchor and positive samples and will be in the same domain as~$\bm{x}$.
\end{itemize}

The combination of these negatives' strategies forces the network to learn to differentiate between images that have different content but similar style (medium negatives) and between images that have similar content but different style (hard negatives), which in the end allows the network to rank images to fulfill its objective, i.e., retrieve images similar in content and style to a given query.

\subsection{Combining Retrieval and Translation}
\label{sec:retrievalTranslation}
As aforementioned, we train our model in three steps: 1) train a image-to-image model without retrieval guidance; 2) train a retrieval system leveraging the model trained in step 1; and 3) fine-tune the image-to-image translation model exploiting the retrieval system. In this section we explain how we use the retrieval model to assist the image translation task in the model trained in step 3.

Given an input image $\bm{x}\in \mathcal{X}_i$ to be translated to domain $\pmb{\mathcal{X}}_j$ we use the retrieval system to find images similar in content to $E_c(\bm{x})$ and with the $\bm{s}_j$ attributes of the target domain. 
To do that we compute the Euclidean distance between the pre-computed embeddings of the query image and the elements in our retrieval set.
We keep the $n$ most similar images $\{\bm{r}_1, \bm{r}_2,\cdots, \bm{r}_{n}\}$ (in our experiments $n=3$) and encode the content of retrieved images using $E_c$: $\{E_c(\bm{r}_1), E_c(\bm{r}_2), \cdots, E_c(\bm{r}_n)\}$.
Then, we concatenate these encodings in the channel dimension. To fuse the features of the retrieved images, we apply a channel-wise convolution. We concatenate that feature map with the content of the input image $E_c(\bm{x})$ in the channel dimension, and again apply a channel wise convolution that fuses the content information of the input image and the retrieved images. We can write the expression of the fused content information as:
\begin{equation}
    \bm{c}_{total}
 = f(E_c(\bm{x}) \oplus f(E_c(\bm{r}_1)\oplus E_c(\bm{r}_2)\oplus\cdots E_c(\bm{r}_n)))
\end{equation}
where $\oplus$ is the concatenation operation and $f$ a channel-wise convolution. 
Finally, we modify the previously defined losses to feed them with $\bm{c}_{total}$ instead of solely with the content of the input image, as we did in step 1.
Thus, for \Cref{eq:1}, \Cref{eq:2}, \Cref{eq:3}, \Cref{eq:4}, \Cref{eq:5} and \Cref{eq:6} we replace the previously defined $\tilde{\bm{c}} = E_c(\bm{x})$ with $\tilde{\bm{c}}= \bm{c}_{total}$.
Different from state-of-the-art methods~\cite{huang2018multimodal,liu2020gmm, gonzalez2018image}, in this paper we feed to the generator not only the content features of the source image, but also those of the retrieved ones. Figure~\ref{fig:RG_pipeline} shows the complete architecture of the retrieval-guided image-to-image translation model.

An attention mechanism is used as a simple yet efficient method to preserve the background pixels while manipulating some visual attributes, which has been proved in state-of-the-art methods~\cite{pumarola2018ganimation,mejjati2018unsupervised,liu2020gmm}. We add a convolutional layer at the end the of decoder $G$ to learn a one-channel attention mask $\bm{M}$. The final generated image $\hat{\bm{x}}$ is obtained through linearly combining the input image $\bm{x}$ and its initial
prediction $\tilde{\bm{x}}$: $\hat{\bm{x}} = \tilde{\bm{x}}*\bm{M} + \bm{x} * (1-\bm{M})$.

\section{Experiments}
We evaluate the proposed retrieval system to ensure it provides valuable information to the retrieval-guided image-to-image translation model. 
We use the CelebFaces Attributes (CelebA) dataset~\cite{liu2015deep} to verify the efficacy of the proposed method. This dataset contains 202,599 face images of celebrities, each one annotated with 40 binary attributes. We crop the initial 178$\times$218 images to 178$\times$178, and then resize them to 128$\times$128. We randomly select 2,000 images as test set and use all remaining images as training data. We follow the setting of the previous methods~\cite{choi2018stargan, liu2020gmm} to construct seven domains using the following attributes: hair color (\textit{black/blond/brown}), gender (\textit{male/female}), and age (\textit{young/old}).

As baseline we use GMM-UNIT~\cite{liu2020gmm}, a multi-domain and multi-modal model that offers to be a unified framework for all the multi-domain image-to-image translation models. As explained before, this model disentangles the content and style latent representations from images, and uses a GMM to model the style latent space.
We did not test the similar MUNIT approach~\cite{huang2018multimodal} which also disentangles content and style, as it is limited to one-to-one domain translations.

\begin{figure*}[t]	
	\renewcommand{\tabcolsep}{1pt}
	\renewcommand{\arraystretch}{0}
	\newcommand{\sizea}{0.105\linewidth}
	\centering
	\footnotesize
	\begin{tabular}{c c cc c cc c cc c cc}
	   \textbf{Query Image} & & \multicolumn{2}{c}{\textbf{Black+Female}} & & \multicolumn{2}{c}{\textbf{Brown+Female}} & & \multicolumn{2}{c}{\textbf{Blond+Male}} & & \multicolumn{2}{c}{\textbf{Blond+Female+Old}} \\  \cmidrule(lr){3-4} \cmidrule(lr){6-7}  \cmidrule(lr){9-10} \cmidrule(lr){12-13}
		\includegraphics[width=\sizea]{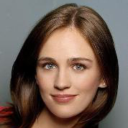} & &
		\includegraphics[width=\sizea]{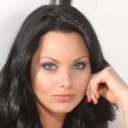} & 
		\includegraphics[width=\sizea]{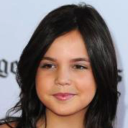} & &
		\includegraphics[width=\sizea]{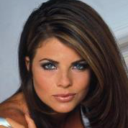} & 
		\includegraphics[width=\sizea]{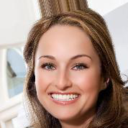} & &
		\includegraphics[width=\sizea]{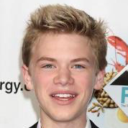} & 
		\includegraphics[width=\sizea]{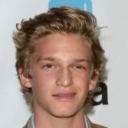} & &
		\includegraphics[width=\sizea]{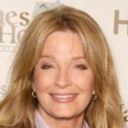} & 
		\includegraphics[width=\sizea]{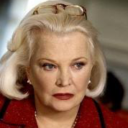}\\ 
		\includegraphics[width=\sizea]{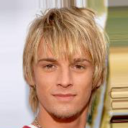} & & 
		\includegraphics[width=\sizea]{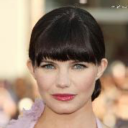} & \includegraphics[width=\sizea]{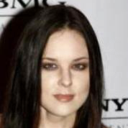} & &
		\includegraphics[width=\sizea]{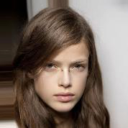} & \includegraphics[width=\sizea]{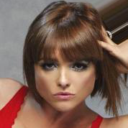} & & 
		\includegraphics[width=\sizea]{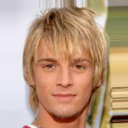} & \includegraphics[width=\sizea]{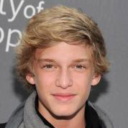} & & 
		\includegraphics[width=\sizea]{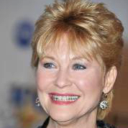} & \includegraphics[width=\sizea]{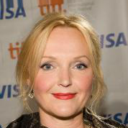}\\
	\end{tabular}
	\vspace{-2pt}
	\caption{Top 2 retrieval examples for different queries. Queries are composed by the content of the query images, on the left, and the attributes resulting by modifying the input image attributes with the ones written on top of each column. 
	}
	\label{fig:retrieval_results}
\end{figure*}

\subsection{Evaluation}

\textbf{Image Retrieval.} We use one query per each test set image with a target domain randomly selected from the testing set, and define several metrics to ensure that the retrieved images are similar in content to the query image and in style to the query attributes. 
Those metrics are based on Precision at 10 (P@10) which measures the similarity of the top 10 retrieval results with the query. 
We use P@10 to measure the following similarities:

\medskip
\noindent \textit{Attributes Similarity.} We evaluate the P@10 of each one of the attributes of the query with the groundtruth attributes of the retrieved images. Then, we average the P@10 of all attributes. 


\medskip
\noindent \textit{Content Similarity.} CelabA groundtruth contains 40 attributes per image, from which only 5 are used to define domains in this work. This metric exploits the additional 35 attributes to provide an insight on how similar the images are in other features aside from the 5 attributes. Therefore, it is an insight of content similarity. It is computed as the former metrics, but 
averaging the P@10 over the 35 unused attributes. Examples of those content attributes are \textit{smiling}, \textit{pointy nose} or \textit{mustache}.

\medskip
\noindent \textit{Average Similarity.} Average between Attributes Similarity and Content Similarity to measure the overall performance, since we aim to train a retrieval system with a good compromise between them. 

\medskip

\noindent\textbf{Image Translation.} We quantitatively evaluate our model through image quality, diversity and domain correctness. Specifically, we use the Inception Score (IS) and the Fréchet Inception Distance (FID)~\cite{NIPS2017_7240} for image quality, the LPIPS~\cite{zhang2018unreasonable} for image diversity and the F1 score for the domain accuracy.

\medskip
\noindent \textit{FID.} Measures the distance in feature space between the real images and the generated ones. We estimate the FID using 10,000 input images vs their transformation in a random target domain.
Lower FID values indicate better quality of synthesized images. 

\medskip
\noindent \textit{IS.} Evaluates the quality of the generated images using their class probabilities computed by a pre-trained network, assuming that strong and varied classification scores indicate higher quality. We compute it using the same transformed images as for FID.

\medskip
\noindent \textit{LPIPS.} The LPIPS is defined as the $\mathcal{L}_2$ distance between the features extracted by a deep learning model of two images and it is a proxy for the perceptual distance between images~\cite{zhang2018unreasonable}.
Consistently with~\cite{huang2018multimodal,lee2018diverse,zhu2017toward}, we randomly select 100 input images and translate them to different domains by generating, for each domain, 10 images and evaluate the average LPIPS distance between the 10 generated images. Then, we average all distances to get the final score. Higher LPIPS distance indicates better diversity among the generated images. 

\medskip
\noindent \textit{ACC.} To evaluate if the generated images have the desired attributes, i.e., they are in the target domain, we train an attribute classifier. We use a pre-trained ResNet-50~\cite{he2016deep}, modifying the last linear layer to have the same number of outputs as our number of attributes. We train the network in a multi-label classification setup using sigmoid activations and a cross-entropy loss. At testing time, we consider an attribute positive if its score is above $0.5$, we evaluate the accuracy (ACC) of each attribute separately and we compute the mean of them.

\section{Results}

\subsection{Image Retrieval}
We begin by evaluating our retrieval system, which is trained to retrieve images similar in content to the query one but with the style of the target domain.
Figure \ref{fig:retrieval_results} shows top two ranked images for different query images and target attributes. 
Note that the retrieved images have the desired style while keeping content similarity (visual appearance not related with the defined attributes) with the query image.

Table \ref{tab:retrieval_scores} shows the quantitative evaluation of our retrieval model trained with different triplet selection strategies.
We observe that the model trained with a combination of easy, medium and hard negatives achieves the best performance.
In particular, the introduction of hard negatives significantly boosts the similarity of the attributes at a small cost of content similarity, resulting in the best average performance. 
This trade-off between content and style similarity is motivated by the fact that hard negatives share the same content as the positives, thus guiding the network to learn image similarities based on style.
These results are consistent for each domain and prove that the proposed triplet selection strategy is crucial to achieve a consistent retrieval performance.
Readers can find additional results in the Supplementary Material.

\begin{table}[ht]
    \caption{Quantitative results on the image retrieval system. For each training strategy we show the P@10 for attributes and content similarity, but also the average between the two.}
    \label{tab:retrieval_scores}
    \begin{tabular}{@{}lrrrr@{}}
    \midrule
     \textbf{Training Triplets} & \textbf{Attributes} & \textbf{Content} & \textbf{Average score} \\
    \midrule
    Easy Negatives & 0.656 & \textbf{0.792} & 0.724 \\
    Medium Negatives & 0.656 & 0.790   &  0.723 \\
    Hard Negatives & 0.797 &  0.786   & 0.791 \\
    All & \textbf{0.830} & 0.775 & \textbf{0.802} \\
    \midrule
    Random & 0.648 & 0.754 & 0.701 \\
    \bottomrule       
\end{tabular}
\end{table}

\begin{figure*}[ht]	
	\renewcommand{\tabcolsep}{2pt}
	\renewcommand{\arraystretch}{0}
	\newcommand{\sizea}{0.11\linewidth}
	\centering
	\footnotesize
	\begin{tabular}{cc ccccc}	   
	\textbf{Input Image} & & \textbf{Black+Female} & \textbf{Blond+Female} & \textbf{Brown+Female}  & \textbf{Blond+Male} & \textbf{Blond+Old+Male} \vspace{2pt} \\ 
	\includegraphics[width=\sizea]{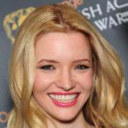} & &
	\includegraphics[width=\sizea]{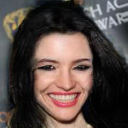} & 
	\includegraphics[width=\sizea]{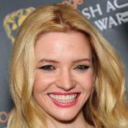} &
	\includegraphics[width=\sizea]{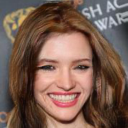} &
	\includegraphics[width=\sizea]{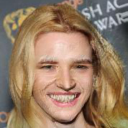} &
	\includegraphics[width=\sizea]{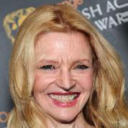} \\ 
	\includegraphics[width=\sizea]{figures/retrieval/124353.png} & &
	\includegraphics[width=\sizea]{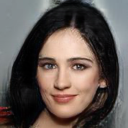} & 
	\includegraphics[width=\sizea]{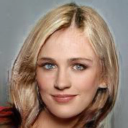} &
	\includegraphics[width=\sizea]{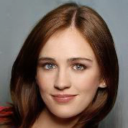} &
	\includegraphics[width=\sizea]{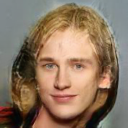} &
	\includegraphics[width=\sizea]{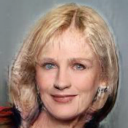} \\ 
	\includegraphics[width=\sizea]{figures/retrieval/055899.png} & & 
	\includegraphics[width=\sizea]{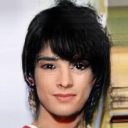} & 
	\includegraphics[width=\sizea]{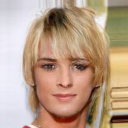} &
	\includegraphics[width=\sizea]{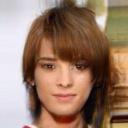} &
	\includegraphics[width=\sizea]{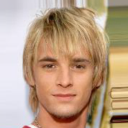} &
	\includegraphics[width=\sizea]{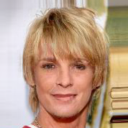} \\ 
	\end{tabular}
	\caption{Qualitative image-to-image transformation results of the RG-UNIT model. 
	}
	\label{fig:img2img_results}
\end{figure*}

\subsection{Image Translation}
\begin{figure}[ht]	
	\renewcommand{\tabcolsep}{1pt}
	\renewcommand{\arraystretch}{0}
	\newcommand{\sizea}{0.185\linewidth}
	\centering
	\footnotesize
	\begin{tabular}{c c cc c cc}	   
	\multirow{2}{*}{\textbf{Input Image}} & & \multicolumn{2}{c}{\textbf{Brown+Female}} & & 
	\multicolumn{2}{c}{\textbf{Blond+Old+Male}} \\ \cmidrule(lr){3-4} \cmidrule(lr){6-7} \vspace{2pt}
	&& \textbf{Attention} & \textbf{Result} && \textbf{Attention} & \textbf{Result} \\
	\includegraphics[width=\sizea]{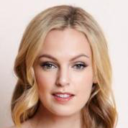} &&
	\includegraphics[width=\sizea]{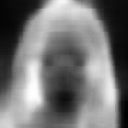} &
	\includegraphics[width=\sizea]{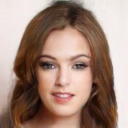} &&
	\includegraphics[width=\sizea]{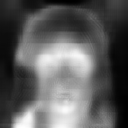} &
	\includegraphics[width=\sizea]{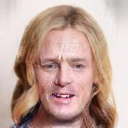} \vspace{2pt} \\
	\multirow{2}{*}{\textbf{Input Image}} & & \multicolumn{2}{c}{\textbf{Black+Female}} & & 
	\multicolumn{2}{c}{\textbf{Blond+Old+Male}} \\ \cmidrule(lr){3-4} \cmidrule(lr){6-7} \vspace{2pt}
	&& \textbf{Attention} & \textbf{Result} && \textbf{Attention} & \textbf{Result} \\
	\includegraphics[width=\sizea]{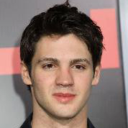} &&
	\includegraphics[width=\sizea]{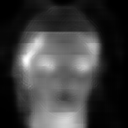} &
	\includegraphics[width=\sizea]{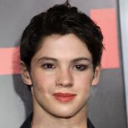} &&
	\includegraphics[width=\sizea]{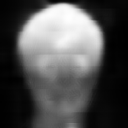} &
	\includegraphics[width=\sizea]{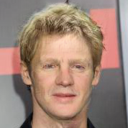} \\
	\end{tabular}
	\caption{Unsupervised-learned attention of the translation.
	}
	\label{fig:attention}
\end{figure}

We now move our attention to the main task: the image-to-image translation. 
Figure \ref{fig:img2img_results} shows some generated results for different input images and target styles. We can see that the proposed RG-UNIT model learns to translate images to multiple domains and that it can model multiple transformations at once (e.g. changing the gender and the age in a single pass). We observe that cross-gender translations are the most challenging transformations as CelebA is an unbalanced dataset, where some attributes are correlated with each other.
As we use an attention mechanism to learn which pixels of the images have to be transformed, the generated results are clean and the transformation affects solely on the attributes to be changed. 
Figure \ref{fig:attention} shows examples of the learned attention masks. 

\begin{figure}[ht]	
	\renewcommand{\tabcolsep}{1pt}
	\renewcommand{\arraystretch}{0}
	\newcommand{\sizea}{0.185\linewidth}
	\centering
	\footnotesize
	\begin{tabular}{cc cccc}
	   \textbf{Input Image} & & \multicolumn{4}{c}{\textbf{Old}} \\  
	   \cmidrule(lr){3-6} 
		\includegraphics[width=\sizea]{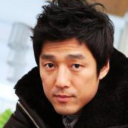} & & 
		\includegraphics[width=\sizea]{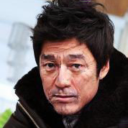} & 
		\includegraphics[width=\sizea]{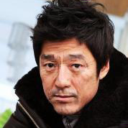} & 
		\includegraphics[width=\sizea]{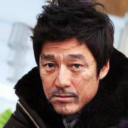} & 
		\includegraphics[width=\sizea]{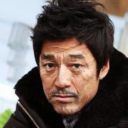} \vspace{2pt} \\ 
	    \textbf{Input Image} & & \multicolumn{4}{c}{\textbf{Blonde}} \\ 
	   \cmidrule(lr){3-6} 
		\includegraphics[width=\sizea]{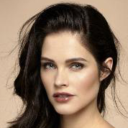} & & 
		\includegraphics[width=\sizea]{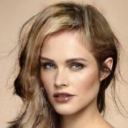} & 
		\includegraphics[width=\sizea]{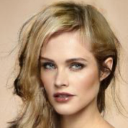} & 
		\includegraphics[width=\sizea]{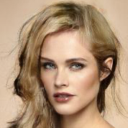} & 
		\includegraphics[width=\sizea]{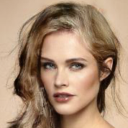} \vspace{2pt} \\ 
	    \textbf{Input Image} & & \multicolumn{4}{c}{\textbf{Male+Black}} \\ 
	    \cmidrule(lr){3-6} 
		\includegraphics[width=\sizea]{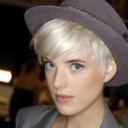} & & 
		\includegraphics[width=\sizea]{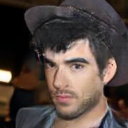} & 
		\includegraphics[width=\sizea]{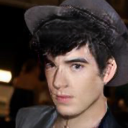} & 
		\includegraphics[width=\sizea]{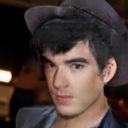} & 
		\includegraphics[width=\sizea]{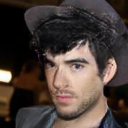} \\ 
	\end{tabular}
	\caption{Different results for a given input image and target transformation. Zoom in for better details.
	}
	\label{Fig:img2img_results_diversity}
\end{figure}

RG-UNIT models the latent style through a GMM, enabling to generate multiple plausible results of the same translation. For an input image and a target style, we can sample multiple latent codes from the latent distribution (the GMM) and generate the images. To verify this claim and that we do not achieve higher quality results than the baseline at the expense of diversity, we generate multiple samples of the same transformation. Thus, given an input image $\bm{x}$ we sample multiple styles $\bm{s}_1, \bm{s}_2, \bm{s}_3$ and generate multiple images $G(Ec(\bm{x}), \bm{s}_1)$, $G(Ec(\bm{x}), \bm{s}_2)$, $G(Ec(\bm{x}), \bm{s}_3)$. \Cref{Fig:img2img_results_diversity} shows different results for an input image and a target transformation. As it can be seen by zooming in, different styles of the faces, beard and hair can be seen among the results.

\begin{figure}[ht]	
	\renewcommand{\tabcolsep}{1pt}
	\renewcommand{\arraystretch}{0.8}
	\newcommand{\sizea}{0.16\columnwidth}
	\footnotesize
	\begin{tabular}{c|cccccccc}
	   Input Image & \multicolumn{2}{c}{Black hair + Female}  &\multicolumn{1}{c}{$\Longleftrightarrow$}  &\multicolumn{2}{c}{Blond hair + Female} \\
	   \includegraphics[width=\sizea]{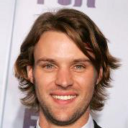} &
	   \includegraphics[width=\sizea]{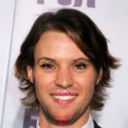} &
	   \includegraphics[width=\sizea]{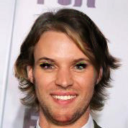} &
	   \includegraphics[width=\sizea]{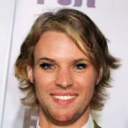} &
	   \includegraphics[width=\sizea]{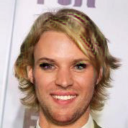} &
	   \includegraphics[width=\sizea]{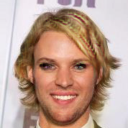} \\
	\end{tabular}
	\caption{Domain interpolation given an input image.}
	\label{Fig:interpolation}
	\vspace{-3mm}
\end{figure}

Moreover, a smooth and continuous latent space allows to interpolate between styles. \Cref{Fig:interpolation} shows an example where we first translate an input image to black hair and female, to then interpolate the latent space and get intermediate results of the transformation to blond hair.

The qualitative results are corroborated also by the quantitative ones. 
Table \ref{tab:translation_scores} shows that RG-UNIT significantly outperforms the framework GMM-UNIT in image quality, measured by both the FID and IS scores. The LPIPS diversity score is similar, corroborating that the retrieval system guidance does not hurt the image generation diversity,

We observe that the classification accuracy for all generated images is high for all the considered models, superior even to the accuracy achieved in real images, but a bit lower for RG-UNIT. 
We hypothesize that this result is a consequence of two main reasons. 
First, CelebA contains significant noise in the labels, which might be alleviated in the generated images thanks to the domain classification loss.
Second, the additional content information provided by the retrieval images might hurt the accuracy of the generated images as the decoder might rely more on the content than on the style to generate images. 

To further prove the contribution of the retrieval system, we test a RG-UNIT version where, instead of using the images top ranked by the retrieval system to assist the training, we retrieve random images. As expected this model gets worse image quality results, but better accuracy (second row of \Cref{tab:translation_scores}).
We refer to the Supplementary Material for additional per-class accuracy. 

Altogether, our results show that the retrieval model guidance is able to boost the image-to-image translation performance.

\begin{table}[ht]
    \caption{Quantitative results for GMM-UNIT, the proposed RG-UNIT, and a model with the same architecture as RG-UNIT but provided with random retrieved images.}
    \label{tab:translation_scores}
    \begin{tabular}{@{}lrrrr@{}}
    \toprule
    \textbf{Model} & \textbf{FID$\downarrow$} & \textbf{IS$\uparrow$} & \textbf{LPIPS$\uparrow$} & \textbf{ACC$\uparrow$} \\
    \midrule  
    GMM-UNIT & 30.04 & 3.051  & \textbf{0.048}  &  97.0             \\
    RG-UNIT - Random Retrieval  & 31.08 & 2.967 & 0.036 &  \textbf{98.2}  \\
    RG-UNIT  & \textbf{27.61} & \textbf{3.160} & 0.042 & 94.72           \\
    \midrule
    Real images  & - & 3.470 & - & 93.1 \\
    \bottomrule
    \end{tabular}
\end{table}

\medskip\noindent\textbf{Exploiting Unlabeled Data.}
The image-to-image translation learning addressed here is unsupervised, meaning that it learns to map one visual domain to another from unpaired images (e.g. two different people with two different visual domains).
However, it still relies on annotations that indicate what features are present in the images. For example, in CelebA we have binary annotations to indicate the gender of a face, the hair color, etc.

One of the main features of the proposed retrieval-guided method is that it can benefit from unlabeled data to train the image translation system.
This can be easily achieved by feeding additional unlabeled data to the retrieval set during the third stage of the training (RG-UNIT training).

To simulate the scenario where we leverage additional unlabeled data, we train the baseline GMM-UNIT and retrieval models with a subset of the training dataset. Then, we train a RG-UNIT model using the same training dataset subset and the whole training data as retrieval set. 
This allows the translation network to benefit from additional unlabeled data through the image retrieval system guidance.

Table \ref{tab:metrics_reduced_dataset} shows image translation metrics for models trained with subsets of the original training data and using all the training data in the retrieval set. Results show that the smaller the dataset is and larger the additional retrieval data is the higher the contribution of the retrieval system to the translation model is. 
This experiment proves that the boost provided by the retrieval guidance is bigger when the training data are scarce, and that the proposed method can benefit from additional unlabeled data.

\begin{table}[ht]
    \caption{Image translation scores of models trained with a subset of the dataset and score shifts provided by the retrieval guidance.}
    \label{tab:metrics_reduced_dataset}
    \begin{tabular}{@{}lc|cccc@{}}
    \toprule  
       & \textbf{Data \%} & \textbf{FID}$\downarrow$ & $\Delta$\textbf{FID} & \textbf{IS}$\uparrow$ &  $\Delta$\textbf{IS} \\
    \midrule  
    GMM-UNIT &  \multirow{2}{*}{25}  & 38.83 & \multirow{2}{*}{4.92$\downarrow$} & 2.648 &  \multirow{2}{*}{0.431$\uparrow$}   \\
    RG-UNIT &    & 33.91 &  & 3.079 &  \\
    \midrule
    GMM-UNIT &  \multirow{2}{*}{50}  & 35.62 &  \multirow{2}{*}{2.90$\downarrow$} & 2.726  & \multirow{2}{*}{0.524$\uparrow$}  \\
    RG-UNIT &   & 32.72 &  & 3.250  & \\
    \midrule
    GMM-UNIT &  \multirow{2}{*}{100}  & 30.04 & \multirow{2}{*}{2.43$\downarrow$} & 3.051 & \multirow{2}{*}{0.109$\uparrow$} \\
    RG-UNIT &   & 27.61 &  & 3.160 & \\ 
    \bottomrule  
    \end{tabular}
\end{table}

\subsection{Ablation Study} We further investigate the role of the number of retrieved images in the performance of RG-UNIT. 
Table \ref{tab:different_retrieved_images} shows that all models improve the baseline performance, but using three retrieved images seems to be the optimal setup. 

\begin{table}[ht]
    \caption{Image translation results for RG-UNIT with different number of retrieved images ($\mathbf{r}$).}
    \label{tab:different_retrieved_images}
    \begin{tabular}{@{}lcrrr@{}}
    \toprule
    \textbf{Model} & $\mathbf{N}$ & \textbf{FID$\downarrow$} & \textbf{IS$\uparrow$} &  \textbf{ACC$\uparrow$} \\
    \midrule  
    RG-UNIT &    1    & 28.45 & 3.053  &    \textbf{97.28}           \\
    RG-UNIT &    3    & \textbf{27.61} & \textbf{3.160} &    94.72           \\
    RG-UNIT &    10    & 28.61 & 3.018 &  96.62            \\
    \midrule
    GMM-UNIT & - & 30.04 & 3.051  &  97.0 \\      
    \bottomrule
    \end{tabular}
\end{table}

\section{Conclusions}
In this paper, we presented a novel approach for unsupervised multi-domain image-to-image translation that uses a retrieval model to improve the quality of the generated images. 
First, we train an image translation model that learns to disentangle content and style from unpaired images.
Then, we leverage these disentangled representations to train an image retrieval system. For that purpose, we design a novel triplet selection strategy exploiting the image translation model, which is proven to be crucial to achieving a superior retrieval performance.
Then, we fine-tune the image-to-image translation model exploiting the retrieval model guidance and learn a better decoder.
Our experiments show that the use of an image retrieval system improves the quality of image-to-image translation, especially when additional unlabelled data is used through the retrieval guidance. In particular, we observe that the improvement in performance increases as the dataset size decreases. As image-to-image translations models require a large amount of data, learning from unlabelled examples is of paramount importance.
We hope that the proposed approach gives rise to further research on exploiting image retrieval systems to improve image translation models.

\bibliographystyle{ACM-Reference-Format}
\bibliography{bibliography}

\clearpage
\appendix

\section{Implementation Details}
\label{suppl:implementation_details}

\subsection{GMM-UNIT}

The first step of our training pipeline is to train a GMM-UNIT~\cite{liu2020gmm} with the CelebFaces Attributes (CelebA) dataset~\cite{liu2015deep}. We follow the setup proposed in previous works~\cite{choi2018stargan, liu2020gmm} to construct seven domains using the following attributes: hair color (\textit{black, blond, brown}), gender (\textit{male/female}), and age (\textit{young/old}). 

We train GMM-UNIT with the default settings: Adam optimizer with $\beta_1=0.5$, $\beta_2=0.999$ and an initial lerning rate of $0.0001$, which is decreased by half every $2e5$ iterations. We use a batch size of 1 and set the loss weights to $\lambda_{s/rec}=10$, $\lambda_{cyc}=10$, $\lambda_{KL}=0.1$ and $\lambda_{iso}=0.1$. We use he domain invariant perceptual loss with weight 0.1 and apply random mirroring to the training images.

\subsection{Retrieval system}
The second step of our training pipeline is to train a retrieval model exploiting the image translation model trained in the first step.
The retrieval model takes as inputs the content and attributes representations extracted by the image translation model. In our experiments with $128x128$ images the content feature map size is $[256,32,32]$ while the attributes representation is a vector with $40$ dimensions, where we have 8 dimensions for each one of the $5$ attributes. A series of convolutional layers are applied to the content features map until it is downsized to $[32,4,4]$. Then, it is flattened and concatenated with the attributes representations, and the output is processed by a 2 layers MLP that generate a representation of $100$ dimensions. All convolutional and linear layers but the last one are followed by ReLu activations and linear normalization. 
The architecture of the retrieval model is described in table \ref{tab:architecture_retrieval}.

The model is trained with a batch size of 32, a margin $m$ of $0.2$ and a learning rate of $0.01$ which is decayed by half every $10K$ iterations. Negatives samples for triplets are selected randomly from one of the categories explained in the main paper. 

\subsection{Generator Fine-tuning}

The third and final step of our training pipeline is to fine-tune the image translation model trained in step one in a setup such that it can exploit the information of the images provided by the retrieval system, which have similar content as the input image but are in the target domain.


We encode the content of each one of the retrieved images using $E_c$ and concatenate those encodings in the channel dimension which. In our experiments with $128x128$ images, that concatenation results in a $[M*256,32,32]$ feature map. To fuse the features of the retrieved images, we apply a convolution that reduces the dimensionality to $[256,32,32]$. Next, we concatenate that feature map with the content of the input image $E_c(I_i)$ in the channel dimension, and again apply a convolution that fuses the content information of the input image and the retrieved images and reduces the dimensionality of the fused feature maps to $[256,32,32]$, the dimensions for content encoding in the image translation model trained in step one. 
The rest of the architecture and training procedure of the image-to-image translation model is kept intact, but using the content that resulted from encoding together the content of the input image and the retrieved images, instead of only the content of the input image.

The training setup and parameters are the same as for GMM-UNIT~\cite{liu2020gmm} but training only the generator while keeping the encoder frozen. This is mandatory since the retrieval system uses the encoder to get the images content information, and updating it would cause the retrieval system to struggle. The architecture of the RG-UNIT model is described in table \ref{tab:architecture}.

\begin{table*}[ht]
	\centering
	\begin{tabular}{@{}lll@{}}
	    \toprule
	    \textbf{Part} & \textbf{Input} $\rightarrow$ \textbf{Output Shape} & \textbf{Layer Information} \\ \midrule
		\multirow{11}{*}{$R$} & ($\frac{h}{4}$, $\frac{w}{4}$, 256) $\rightarrow$ ($\frac{h}{4}$, $\frac{w}{4}$, 128) & CONV-(N128, K3x3, S1, P1), IN, ReLU \\
		& ($\frac{h}{4}$, $\frac{w}{4}$, 128) $\rightarrow$ ($\frac{h}{8}$, $\frac{w}{8}$, 128) & CONV-(N128, K4x4, S2, P1), IN, ReLU \\
        & ($\frac{h}{8}$, $\frac{w}{8}$, 128) $\rightarrow$ ($\frac{h}{8}$, $\frac{w}{8}$, 64) & CONV-(N64, K3x3, S1, P1), IN, ReLU \\
        & ($\frac{h}{8}$, $\frac{w}{8}$, 64) $\rightarrow$ ($\frac{h}{16}$, $\frac{w}{16}$, 64) & CONV-(N64, K4x4, S2, P1), IN, ReLU \\
        & ($\frac{h}{16}$, $\frac{w}{16}$, 64) $\rightarrow$ ($\frac{h}{16}$, $\frac{w}{16}$, 32) & CONV-(N32, K3x3, S1, P1), IN, ReLU \\
        & ($\frac{h}{16}$, $\frac{w}{16}$, 32) $\rightarrow$ ($\frac{h}{32}$, $\frac{w}{32}$, 32) & CONV-(N32, K4x4, S2, P1), IN, ReLU \\
        \cmidrule{2-3}
        & ($\frac{h}{32}$, $\frac{w}{32}$, 32) $\rightarrow$ ($\frac{h}{32}$ $\frac{w}{32}$ 32,) & FLATTEN \\
        & ($\frac{h}{32}$ $\frac{w}{32}$ 32,) $\rightarrow$ ($\frac{h}{32}$ $\frac{w}{32}$ 32 + $Zn$,) & CONCAT-($Zn$) \\
        \cmidrule{2-3}
        & ($\frac{h}{32}$ $\frac{w}{32}$ 32 + $Zn$,)  $\rightarrow$ (512,) & FC-(512), ReLU \\
        & (512,) $\rightarrow$ (512,) & FC-(512), ReLU, Dropout-(0.1) \\
        & (512,) $\rightarrow$ (100,) & FC-(100) \\

		\bottomrule
	\end{tabular}
	\vspace{1mm}
	\caption{Retrieval Network architecture. We use the following notation: $Z$: the dimension of attribute vector, $n$: the number of the domains, N: the number of output channels, K: kernel size, S: stride size, P: padding size, CONV: a convolutional layer, FC: fully connected layer, FLATTEN: flattens a d-dimensional tensor into a 1-dimensional tensor, CONCAT-(c): concatenates two 1-dimensional tensors, c and the input. 
	}
	\label{tab:architecture_retrieval}
\end{table*}

\subsection{Source code}
We release the source code of our model at \url{https://github.com/yhlleo/RG-UNIT}.

\section{Additional Quantitative Results}

\subsection{Retrieved Images Attributes Similarity}

Table~\ref{tab:retrieval_scores_per_attribute} shows the P@10 of each one of the attributes for retrieval models trained with different negatives strategies. As seen in the paper, it shows that the negatives selection strategy proposed in this work is crucial to improve the retrieval system performance for all the attributes.

\begin{table}[ht]
    \caption{Image retrieval P@10 per attribute metric (and average between attributes). Note that this metric does not measure content similarity. We show results of the model we use to boost the image-to-image model (All), a model trained only with easy negatives (Easy Negatives), only easy and medium negatives (Medium Negatives) and easy, medium and hard negatives (Hard Negatives). Random refers to the metrics obtained retrieving random images.}
    \label{tab:retrieval_scores_per_attribute}
    \begin{tabular}{@{}lrrrrrr@{}}
    \midrule
    \multicolumn{1}{c}{\textbf{}} & \multicolumn{6}{c}{\textbf{P@10}}                        \\
    \toprule
     & \textbf{BA} & \textbf{BN} & \textbf{BW} & \textbf{G} & \textbf{A} & \textbf{Avg} \\
    \midrule
    Easy Negatives & 0.636 & 0.778 & 0.666 & 0.530 & 0.671 & 0.656 \\
    Medium Negatives  & 0.640 & 0.779 & 0.674 & 0.521 & 0.665 & 0.655 \\
    Hard Negatives  & 0.759 & 0.902 & \textbf{0.756} & 0.796 & 0.770 & 0.797 \\
    All              & \textbf{0.810} & \textbf{0.919} & 0.737 & \textbf{0.841} & \textbf{0.841} & \textbf{0.830} \\
    \midrule
    Random Retrieval  & 0.640 & 0.768 & 0.676 & 0.509 & 0.647 & 0.648  \\
    \bottomrule       
\end{tabular}
\end{table}

\subsection{Attribute Classification}

Table~\ref{tab:translation_acc} shows the attributes classifier accuracy for each one of the attributes on the images generated the GMM-UNIT model, our retrieval-guided RG-UNIT model, a model guided with random images (RG-UNIT RR) and real images.

\begin{table}[ht]
    \caption{Accuracy of the attributes classifier evaluated on the real images from CelaB dataset, on the images generated by GMM-UNIT, by the retrieval-guided model RG-UNIT, and by a model guided with random retrieval images (RG-UNIT RR). BA refers to Black Hair attribute, BN to Blond Hair, BW to Brown Hair, G to Gender, A to Age and Avg to the average scores between them.}
    \label{tab:translation_acc}
    \begin{tabular}{@{}lrrrrrr@{}}
    \toprule
    & \textbf{Avg} & \textbf{BA} & \textbf{BN} & \textbf{BW} & \textbf{G} & \textbf{A}\\
    \midrule
    GMM-UNIT & 97.0 &  97.7 &	\textbf{98.5} & 93.2 & 97.3 & 98.3   \\
    RG-UNIT - RR  & \textbf{98.2} &  \textbf{98.3} &	96.1 &	\textbf{99.5} &   \textbf{98.9} &	\textbf{98.4} \\
    RG-UNIT  & 94.7 &  95.0 &	96.1 &	89.4 &	98.1 &	95.0 \\
        \midrule
    Real Images  & 93.1  & 94.0 & 98.3 & 87.0 &  94.4 & 91.8   \\
    \bottomrule       
    \end{tabular}
\end{table}

\newpage\section{Additional Qualitative Results}

\begin{figure*}[ht]	
	\renewcommand{\tabcolsep}{1pt}
	\renewcommand{\arraystretch}{0}
	\newcommand{\sizea}{0.10\linewidth}
	\centering
	\footnotesize
	\begin{tabular}{c c cc c cc c cc c cc}
	   \textbf{Query Image} & & \multicolumn{2}{c}{\textbf{Brown+Male}} & & \multicolumn{2}{c}{\textbf{Blond+Female}} & & \multicolumn{2}{c}{\textbf{Blond+Male+Young}} & & \multicolumn{2}{c}{\textbf{Black+Female+Old}} \\  \cmidrule(lr){3-4} \cmidrule(lr){6-7}  \cmidrule(lr){9-10} \cmidrule(lr){12-13}
		\includegraphics[width=\sizea]{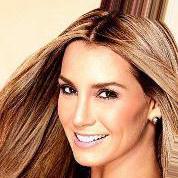} & & 
		\includegraphics[width=\sizea]{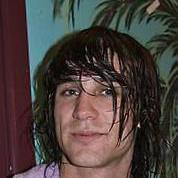} & 
		\includegraphics[width=\sizea]{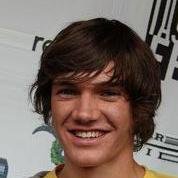} & &  
		\includegraphics[width=\sizea]{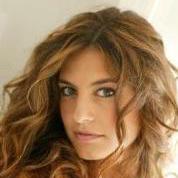} & 
		\includegraphics[width=\sizea]{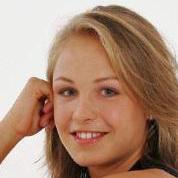} & & 
		\includegraphics[width=\sizea]{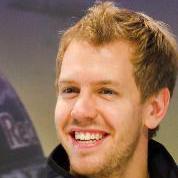} & 
		\includegraphics[width=\sizea]{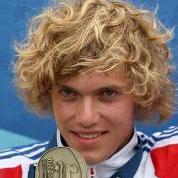} & & 
		\includegraphics[width=\sizea]{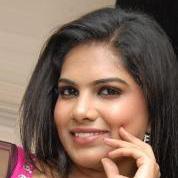} & 
		\includegraphics[width=\sizea]{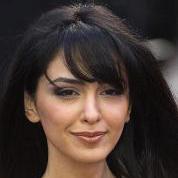} \\
		\includegraphics[width=\sizea]{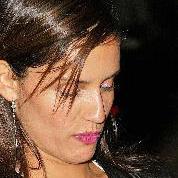} & & 
		\includegraphics[width=\sizea]{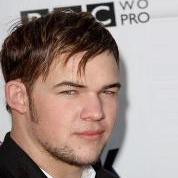} & 
		\includegraphics[width=\sizea]{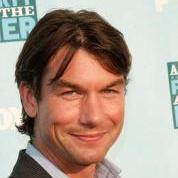} & &  
		\includegraphics[width=\sizea]{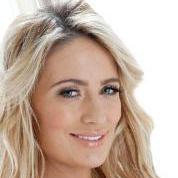} & 
		\includegraphics[width=\sizea]{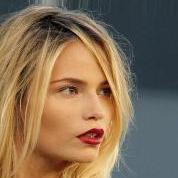} & & 
		\includegraphics[width=\sizea]{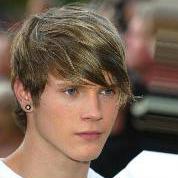} & 
		\includegraphics[width=\sizea]{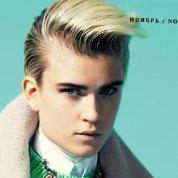} & & 
		\includegraphics[width=\sizea]{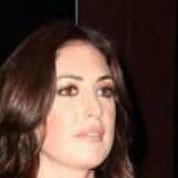} & 
		\includegraphics[width=\sizea]{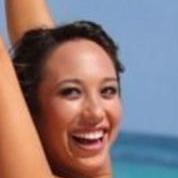} \\
		\includegraphics[width=\sizea]{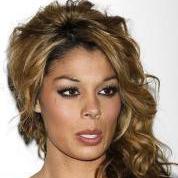} & & 
		\includegraphics[width=\sizea]{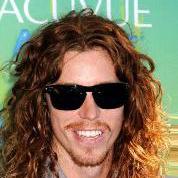} & 
		\includegraphics[width=\sizea]{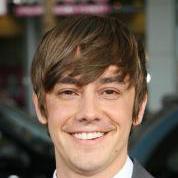} & &  
		\includegraphics[width=\sizea]{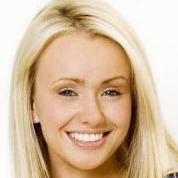} & 
		\includegraphics[width=\sizea]{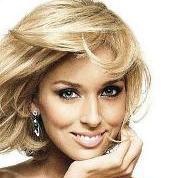} & & 
		\includegraphics[width=\sizea]{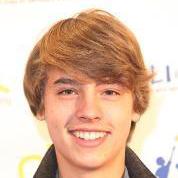} & 
		\includegraphics[width=\sizea]{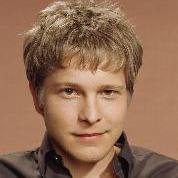} & & 
		\includegraphics[width=\sizea]{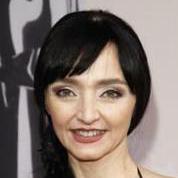} & 
		\includegraphics[width=\sizea]{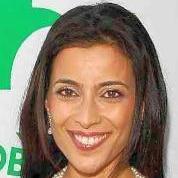} \\
		\includegraphics[width=\sizea]{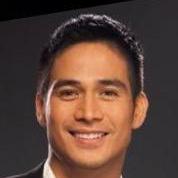} & & 
		\includegraphics[width=\sizea]{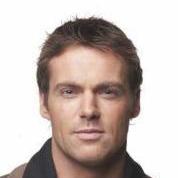} & 
		\includegraphics[width=\sizea]{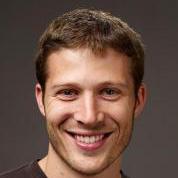} & &  
		\includegraphics[width=\sizea]{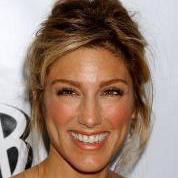} & 
		\includegraphics[width=\sizea]{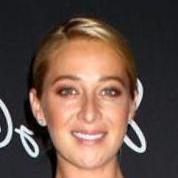} & & 
		\includegraphics[width=\sizea]{figures/app_ret/000352.jpg} & 
		\includegraphics[width=\sizea]{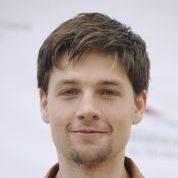} & & 
		\includegraphics[width=\sizea]{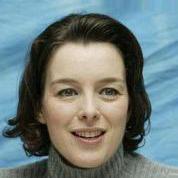} & 
		\includegraphics[width=\sizea]{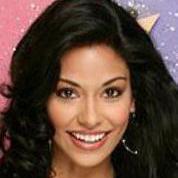} \\
		\includegraphics[width=\sizea]{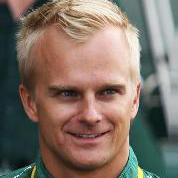} & & 
		\includegraphics[width=\sizea]{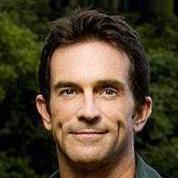} & 
		\includegraphics[width=\sizea]{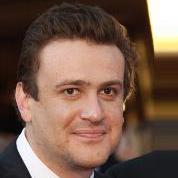} & &  
		\includegraphics[width=\sizea]{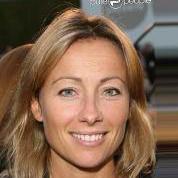} & 
		\includegraphics[width=\sizea]{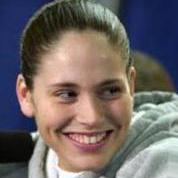} & & 
		\includegraphics[width=\sizea]{figures/app_ret/000116.jpg} & 
		\includegraphics[width=\sizea]{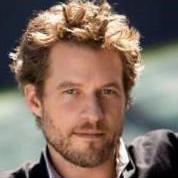} & & 
		\includegraphics[width=\sizea]{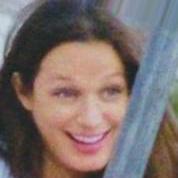} & 
		\includegraphics[width=\sizea]{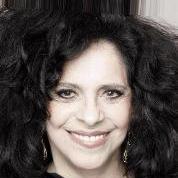} \\
		\includegraphics[width=\sizea]{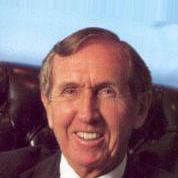} & & 
		\includegraphics[width=\sizea]{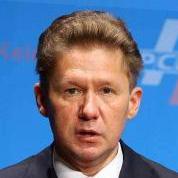} & 
		\includegraphics[width=\sizea]{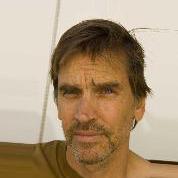} & &  
		\includegraphics[width=\sizea]{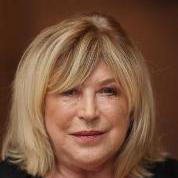} & 
		\includegraphics[width=\sizea]{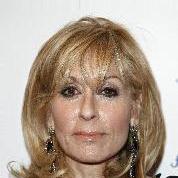} & & 
		\includegraphics[width=\sizea]{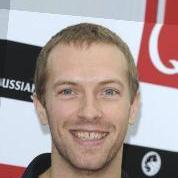} & 
		\includegraphics[width=\sizea]{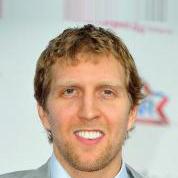} & & 
		\includegraphics[width=\sizea]{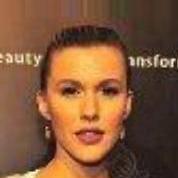} & 
		\includegraphics[width=\sizea]{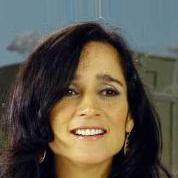} \\
		\includegraphics[width=\sizea]{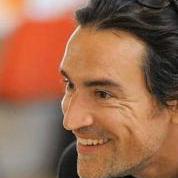} & & 
		\includegraphics[width=\sizea]{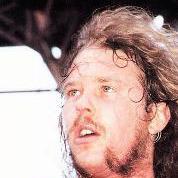} & 
		\includegraphics[width=\sizea]{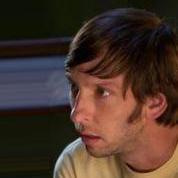} & &  
		\includegraphics[width=\sizea]{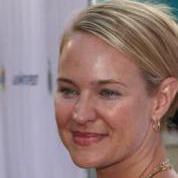} & 
		\includegraphics[width=\sizea]{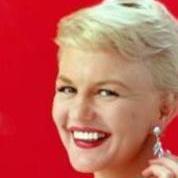} & & 
		\includegraphics[width=\sizea]{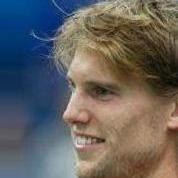} & 
		\includegraphics[width=\sizea]{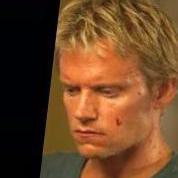} & & 
		\includegraphics[width=\sizea]{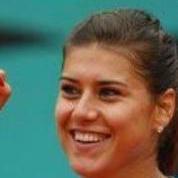} & 
		\includegraphics[width=\sizea]{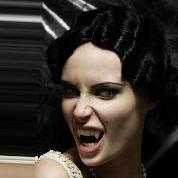} \\
		\includegraphics[width=\sizea]{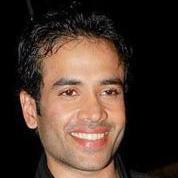} & & 
		\includegraphics[width=\sizea]{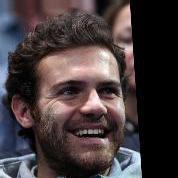} & 
		\includegraphics[width=\sizea]{figures/app_ret/001230.jpg} & &  
		\includegraphics[width=\sizea]{figures/app_ret/000634.jpg} & 
		\includegraphics[width=\sizea]{figures/app_ret/001304.jpg} & & 
		\includegraphics[width=\sizea]{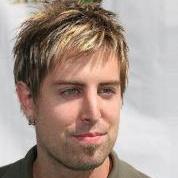} & 
		\includegraphics[width=\sizea]{figures/app_ret/000723.jpg} & & 
		\includegraphics[width=\sizea]{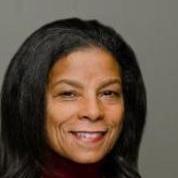} & 
		\includegraphics[width=\sizea]{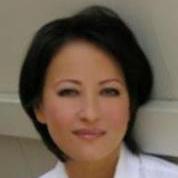} \\
		\includegraphics[width=\sizea]{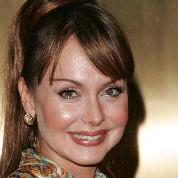} & & 
		\includegraphics[width=\sizea]{figures/app_ret/000689.jpg} & 
		\includegraphics[width=\sizea]{figures/app_ret/000842.jpg} & &  
		\includegraphics[width=\sizea]{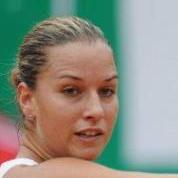} & 
		\includegraphics[width=\sizea]{figures/app_ret/000576.jpg} & & 
		\includegraphics[width=\sizea]{figures/app_ret/000459.jpg} & 
		\includegraphics[width=\sizea]{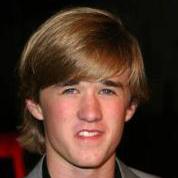} & & 
		\includegraphics[width=\sizea]{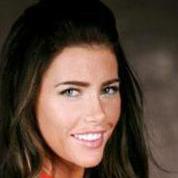} & 
		\includegraphics[width=\sizea]{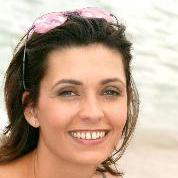} \\
		\includegraphics[width=\sizea]{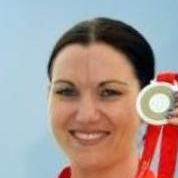} & & 
		\includegraphics[width=\sizea]{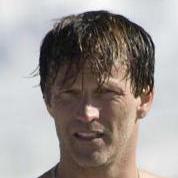} & 
		\includegraphics[width=\sizea]{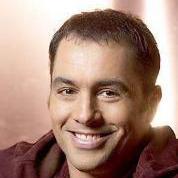} & &  
		\includegraphics[width=\sizea]{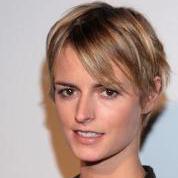} & 
		\includegraphics[width=\sizea]{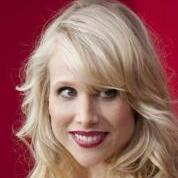} & & 
		\includegraphics[width=\sizea]{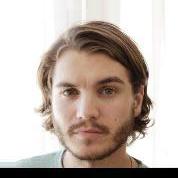} & 
		\includegraphics[width=\sizea]{figures/app_ret/000791.jpg} & & 
		\includegraphics[width=\sizea]{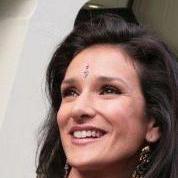} & 
		\includegraphics[width=\sizea]{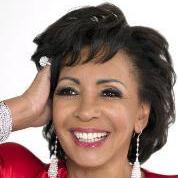} \\
		\includegraphics[width=\sizea]{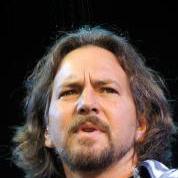} & & 
		\includegraphics[width=\sizea]{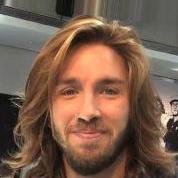} & 
		\includegraphics[width=\sizea]{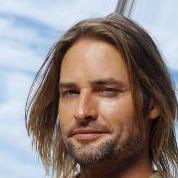} & &  
		\includegraphics[width=\sizea]{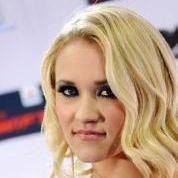} & 
		\includegraphics[width=\sizea]{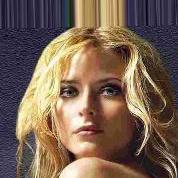} & & 
		\includegraphics[width=\sizea]{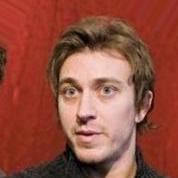} & 
		\includegraphics[width=\sizea]{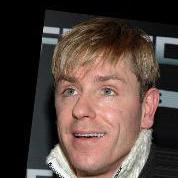} & & 
		\includegraphics[width=\sizea]{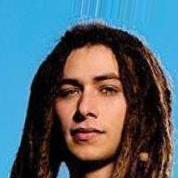} & 
		\includegraphics[width=\sizea]{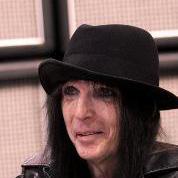} \\
	\end{tabular}
	\vspace{-2pt}
	\caption{Top 2 retrieval examples of the validation set for different queries. Queries are composed by the content of the query images, on the left, and the attributes resulting on modifying the input image attributes with the ones written on top of each column.
	}
	\label{Fig:additional_retrieval_results}
\end{figure*}

Figure~\ref{Fig:additional_retrieval_results} shows additional image retrieval results of the proposed retrieval system. Note how the retrieval system is able to find images with the query attributes but keeping the query image content, such as the mouth gesture or the head pose.

\begin{figure*}[ht]
	\footnotesize
	\begin{tabular}{c}
	   \includegraphics[width=0.68\linewidth]{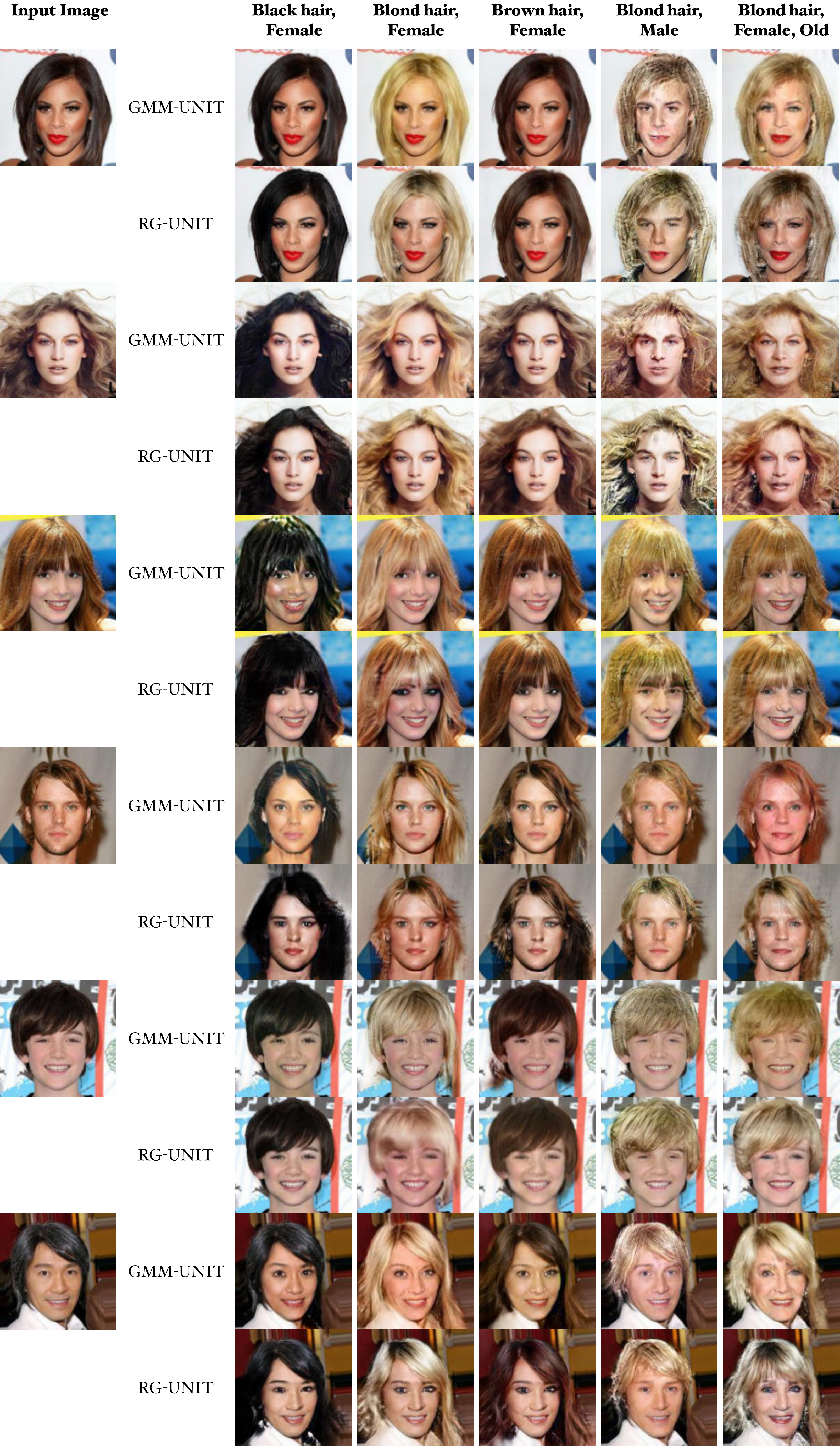}\\
	\end{tabular}
	\caption{Visual comparisons between GMM-UNIT and RG-UNIT.}
	\label{Fig:appendix-rgunit}
\end{figure*}

Figure~\ref{Fig:appendix-rgunit} shows an image translation results comparison between the baseline GMM-UNIT and the proposed RG-UNIT. It evidences how the retrieval guidance helps generating more realistic images.

\begin{figure*}[ht]
	\footnotesize
	\begin{tabular}{c}
	   \includegraphics[width=0.6\linewidth]{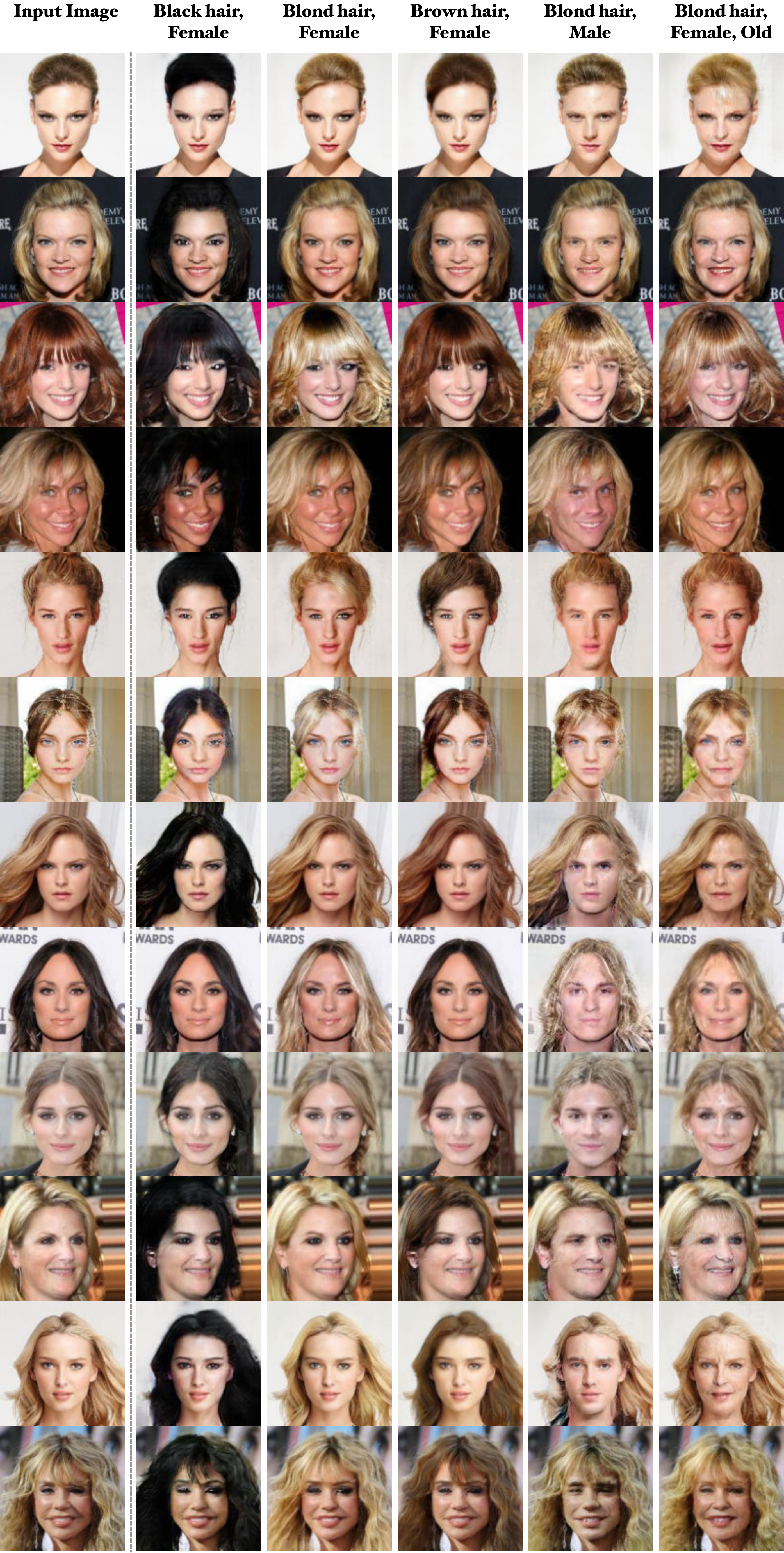}\\
	\end{tabular}
	\caption{More female visual results of our method on CelebA dataset.}
	\label{Fig:appendix-rgunit1}
\end{figure*}

\begin{figure*}[ht]
	\footnotesize
	\begin{tabular}{c}
	   \includegraphics[width=0.6\linewidth]{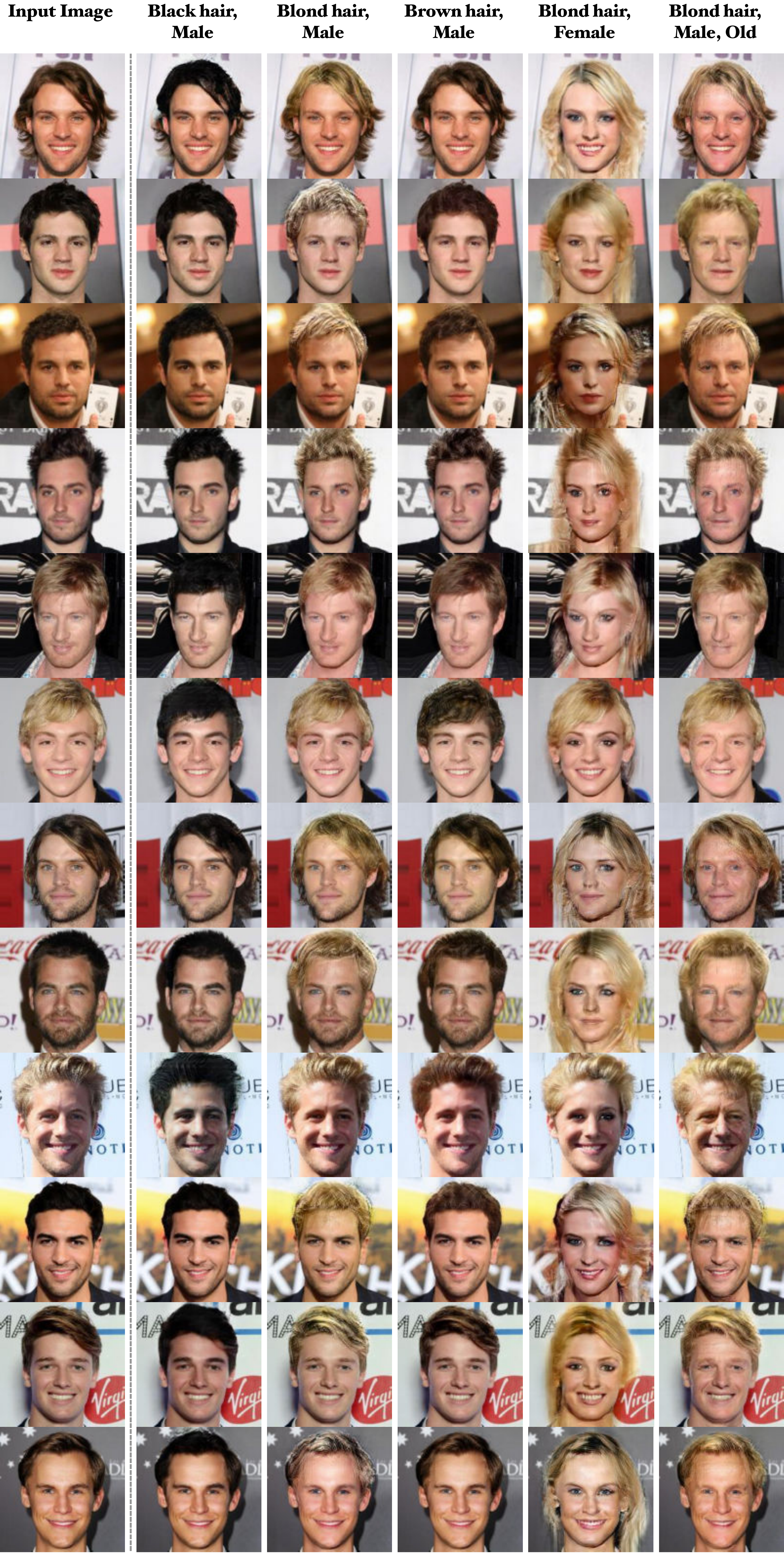}\\
	\end{tabular}
	\caption{More male visual results of our method on CelebA dataset.}
	\label{Fig:appendix-rgunit2}
\end{figure*}

Figures~\ref{Fig:appendix-rgunit1},~\ref{Fig:appendix-rgunit2} show additional image translation results of the proposed RG-UNIT model.

\begin{table*}[ht]
	\centering
	\begin{tabular}{@{}lll@{}}
	    \toprule
	    \textbf{Part} & \textbf{Input} $\rightarrow$ \textbf{Output Shape} & \textbf{Layer Information} \\ \midrule
		\multirow{7}{*}{$E_c$} & ($h$, $w$, 3) $\rightarrow$ ($h$, $w$, 64) & CONV-(N64, K7x7, S1, P3), IN, ReLU \\
		& ($h$, $w$, 64) $\rightarrow$ ($\frac{h}{2}$, $\frac{w}{2}$, 128) & CONV-(N128, K4x4, S2, P1), IN, ReLU \\
		& ($\frac{h}{2}$, $\frac{w}{2}$, 128) $\rightarrow$ ($\frac{h}{4}$, $\frac{w}{4}$, 256) & CONV-(N256, K4x4, S2, P1), IN, ReLU \\ 
		& ($\frac{h}{4}$, $\frac{w}{4}$, 256) $\rightarrow$ ($\frac{h}{4}$, $\frac{w}{4}$, 256) & Residual Block: CONV-(N256, K3x3, S1, P1), IN, ReLU \\ 
		& ($\frac{h}{4}$, $\frac{w}{4}$, 256) $\rightarrow$ ($\frac{h}{4}$, $\frac{w}{4}$, 256) & Residual Block: CONV-(N256, K3x3, S1, P1), IN, ReLU \\ 
		& ($\frac{h}{4}$, $\frac{w}{4}$, 256) $\rightarrow$ ($\frac{h}{4}$, $\frac{w}{4}$, 256) & Residual Block: CONV-(N256, K3x3, S1, P1), IN, ReLU \\ 
		& ($\frac{h}{4}$, $\frac{w}{4}$, 256) $\rightarrow$ ($\frac{h}{4}$, $\frac{w}{4}$, 256) & Residual Block: CONV-(N256, K3x3, S1, P1), IN, ReLU \\ \midrule
		\multirow{2}{*}{$f$} & ($\frac{h}{4}$, $\frac{h}{4}$, 256xM) $\rightarrow$ ($\frac{h}{4}$, $\frac{h}{4}$, 256) & CONV-(N256, K3x3, S1, P1), ReLU \\ 
		& ($\frac{h}{4}$, $\frac{h}{4}$, 256x2) $\rightarrow$ ($\frac{h}{4}$, $\frac{h}{4}$, 256) & CONV-(N256, K3x3, S1, P1), ReLU \\\midrule
		\multirow{8}{*}{$E_z$} & ($h$, $w$, 3) $\rightarrow$ ($h$, $w$, 64) & CONV-(N64, K7x7, S1, P3), ReLU \\
		& ($h$, $w$, 64) $\rightarrow$ ($\frac{h}{2}$, $\frac{w}{2}$, 128) & CONV-(N128, K4x4, S2, P1), ReLU \\
		& ($\frac{h}{2}$, $\frac{w}{2}$, 128) $\rightarrow$ ($\frac{h}{4}$, $\frac{w}{4}$, 256) & CONV-(N256, K4x4, S2, P1), ReLU \\ 
		& ($\frac{h}{4}$, $\frac{w}{4}$, 256) $\rightarrow$ ($\frac{h}{8}$, $\frac{w}{8}$, 256) & CONV-(N256, K4x4, S2, P1), ReLU \\
		& ($\frac{h}{8}$, $\frac{w}{8}$, 256) $\rightarrow$ ($\frac{h}{16}$, $\frac{w}{16}$, 256) & CONV-(N256, K4x4, S2, P1), ReLU \\
		& ($\frac{h}{8}$, $\frac{w}{8}$, 256) $\rightarrow$ (1, 1, 256) & GAP \\ \cmidrule{2-3}
		& (256,) $\rightarrow$ ($CZ$,)  & FC-(N$CZ$) \\
		& (256,) $\rightarrow$ ($CZ$,)  & FC-(N$CZ$) \\
		\midrule
		\multirow{9}{*}{$G$} & ($\frac{h}{4}$, $\frac{w}{4}$, 256) $\rightarrow$ ($\frac{h}{4}$, $\frac{w}{4}$, 256) & Residual Block: CONV-(N256, K3x3, S1, P1), AdaIN, ReLU \\
		& ($\frac{h}{4}$, $\frac{w}{4}$, 256) $\rightarrow$ ($\frac{h}{4}$, $\frac{w}{4}$, 256) & Residual Block: CONV-(N256, K3x3, S1, P1), AdaIN, ReLU \\
		& ($\frac{h}{4}$, $\frac{w}{4}$, 256) $\rightarrow$ ($\frac{h}{4}$, $\frac{w}{4}$, 256) & Residual Block: CONV-(N256, K3x3, S1, P1), AdaIN, ReLU \\
		& ($\frac{h}{4}$, $\frac{w}{4}$, 256) $\rightarrow$ ($\frac{h}{4}$, $\frac{w}{4}$, 256) & Residual Block: CONV-(N256, K3x3, S1, P1), AdaIN, ReLU \\
		& ($\frac{h}{4}$, $\frac{w}{4}$, 256) $\rightarrow$ ($\frac{h}{2}$, $\frac{w}{2}$, 128) & UPCONV-(N128, K5x5, S1, P2), LN, ReLU \\ 
		& ($\frac{h}{2}$, $\frac{w}{2}$, 128) $\rightarrow$ ($h$, $w$, 64) & UPCONV-(N64, K5x5, S1, P2), LN, ReLU \\ 
		& (${h}$, ${w}$, 64) $\rightarrow$ (${h}$, ${w}$, 3) & CONV-(N3, K7x7, S1, P3), Tanh \\ 
		& (${h}$, ${w}$, 64) $\rightarrow$ (${h}$, ${w}$, 1) & CONV-(N3, K7x7, S1, P3), Sigmoid \\ \midrule
		\multirow{7}{*}{$D$} & ($h$, $w$, 3) $\rightarrow$ ($\frac{h}{2}$, $\frac{w}{2}$, 64) & CONV-(N64, K4x4, S2, P1), Leaky ReLU \\
		& ($\frac{h}{2}$, $\frac{w}{2}$, 64) $\rightarrow$ ($\frac{h}{4}$, $\frac{w}{4}$, 128) & CONV-(N128, K4x4, S2, P1), Leaky ReLU \\
		& ($\frac{h}{4}$, $\frac{w}{4}$, 128) $\rightarrow$ ($\frac{h}{8}$, $\frac{w}{8}$, 256) & CONV-(N256, K4x4, S2, P1), Leaky ReLU \\
		& ($\frac{h}{8}$, $\frac{w}{8}$, 256) $\rightarrow$ ($\frac{h}{16}$, $\frac{w}{16}$, 512) & CONV-(N512, K4x4, S2, P1), Leaky ReLU \\  \cmidrule{2-3}
		& ($\frac{h}{16}$, $\frac{w}{16}$, 512) $\rightarrow$ ($\frac{h}{16}$, $\frac{w}{16}$, 1) & CONV-(N1, K1x1, S1, P0) \\ 
		& ($\frac{h}{16}$, $\frac{w}{16}$, 512) $\rightarrow$ (1, 1, $n$) & CONV-(N$n$, K$\frac{h}{16}$x$\frac{w}{16}$, S1, P0) \\ 
		\bottomrule
	\end{tabular}
	\caption{Image-to-image network architecture. We use the following notations: $Z$: the dimension of attribute vector, $n$: the number of attributes, N: the number of output channels, K: kernel size, S: stride size, P: padding size, CONV: a convolutional layer, GAP: a global average pooling layer, UPCONV: a 2$\times$ bilinear upsampling layer followed by a convolutional layer, FC: fully connected layer, M: the number of retrieved images. We set $C=8$.}
	\label{tab:architecture}
\end{table*}

\end{document}